\PassOptionsToPackage{table}{xcolor}
\documentclass[preprint]{costa_preprint}

\usepackage{amsthm}
\usepackage{amssymb}
\usepackage{mathtools}
\usepackage{hyperref}
\usepackage[nameinlink,capitalize]{cleveref}
\usepackage[normalem]{ulem} % for command \sout "strike out"
\usepackage{mathtools}

\usepackage[utf8]{inputenc} % allow utf-8 input
\usepackage[T1]{fontenc}    % use 8-bit T1 fonts
\usepackage{hyperref}       % hyperlinks
\usepackage{url}            % simple URL typesetting
\usepackage{booktabs}       % professional-quality tables
\usepackage{amsfonts}       % blackboard math symbols
\usepackage{nicefrac}       % compact symbols for 1/2, etc.
\usepackage{microtype}      % microtypography
\usepackage{color, colortbl}
\definecolor{Gray}{gray}{0.9}
\newcolumntype{g}{>{\columncolor{Gray}}c}

\usepackage{graphicx,wrapfig}
\usepackage{float}
\usepackage{algorithm}
\usepackage[noend]{algpseudocode}
\usepackage{caption}
\usepackage{subcaption}
\usepackage{multirow}
\usepackage{amsfonts}
\usepackage{tabularx}
\usepackage{url}
\usepackage{enumitem}
\usepackage{amsthm}
\usepackage{thmtools}
\usepackage{thm-restate}

\theoremstyle{definition}

\theoremstyle{remark}

\usepackage{listings}
\usepackage{tcolorbox}
\tcbuselibrary{breakable,skins}

\newtcolorbox{takeawaybox}[2][]{
    enhanced,
    boxsep = 2pt,
    left = 2pt,
    right = 2pt,
    top = 2pt,
    bottom = 2pt,
    #1
}

\newtcolorbox{promptbox}[1][]{
  breakable,
  enhanced,
  colback=gray!5,
  colframe=gray!50,
  boxrule=0.5pt,
  arc=2pt,
  left=6pt, right=6pt, top=6pt, bottom=6pt,
  fonttitle=\bfseries,
  title=#1,
}

\lstdefinestyle{promptstyle}{
  basicstyle=\ttfamily\scriptsize,
  backgroundcolor=\color{gray!8},
  frame=single,
  rulecolor=\color{gray!40},
  breaklines=true,
  breakatwhitespace=true,
  columns=fullflexible,
  keepspaces=true,
  showstringspaces=false,
  xleftmargin=4pt,
  xrightmargin=4pt,
  aboveskip=8pt,
  belowskip=8pt,
  escapeinside={(*}{*)},
}

\definecolor{revisionblue}{HTML}{4C6DA1}
\newcommand{\rev}[1]{\textcolor{revisionblue}{#1}}
\newcommand{\vx}{{\mathbf{x}}}

\DeclareMathOperator*{\argmax}{arg\,max}

\newcommand{\bc}{\begin{center}}
\newcommand{\ec}{\end{center}}

\newcommand{\bdm}{\begin{displaymath}}
\newcommand{\edm}{\end{displaymath}}

\newcommand{\beq}{\begin{equation}}
\newcommand{\eeq}{\end{equation}}

\newcommand{\bfl}{\begin{flushleft}}
\newcommand{\efl}{\end{flushleft}}

\newcommand{\bt}{\begin{tabbing}}
\newcommand{\et}{\end{tabbing}}

\newcommand{\beqn}{\begin{align}}
\newcommand{\eeqn}{\end{align}}

\newcommand{\beqs}{\begin{align*}} % no equation numbers
\newcommand{\eeqs}{\end{align*}}  % no equation numbers

\usepackage{proof-at-the-end}

\ifcostareview
    \usepackage{lineno}
\fi
\ifcostapreprint
    \renewcommand{\rev}[1]{#1}
\fi
\hypersetup{colorlinks=true, citecolor=nusblue, linkcolor=nusblue, urlcolor=nusblue}

\title{Revisiting LLMs as Context Engineers}
\author[1]{Junrui~Zhang}
\author[2]{Zeyu~Xu}
\author[1]{Runchuan~Zhu}
\author[2]{Jie~Peng}
\author[2]{Long~Zhang}
\author[3]{Zixin~Ding}
\author[4]{Jiaxin~Lin}
\author[1]{Kai~Wang}
\author[2]{Jianmin~Ji}
\author[1]{Junyuan~Hong}

\affiliation[1]{National University of Singapore}
\affiliation[2]{University of Science and Technology of China}
\affiliation[3]{University~of~Chicago}
\affiliation[4]{Independent~Researcher}
\correspondence{Junyuan Hong at \email{jyhong@nus.edu.sg}}
\metadata[Code]{\url{https://github.com/costa-nus/LCEBench}}
\logo{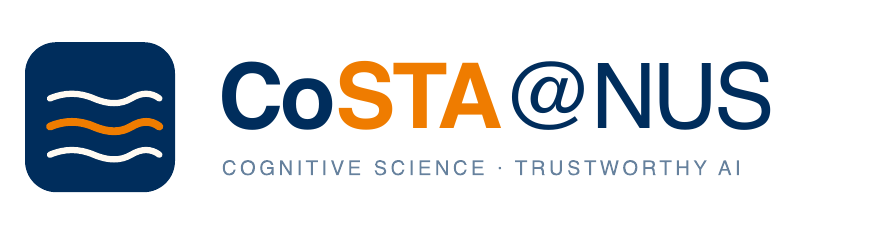}

\begin{document}

\ifcostareview
    \linenumbers
\fi

\abstract{
    As large language models (LLMs) are used in more and more new tasks, context engineering becomes a fundamental method to efficiently steer LLMs by constructing instruction prompts or metadata without intensive training.
    A growing number of LLM context engineering (LCE) approaches automate the task by casting the LLM itself as the engineer that draws generalizable instructions or experiences from offline trials as context.
    Although these methods resemble how humans learn from experiences, it remains unclear \textbf{when and how} LLMs can engineer effective context from data.
    To answer the question, we first establish a benchmark composed of datasets with different reasoning patterns.
    We compared different LCE methods with different LLM families in 8B size, showing that there is no single winner for all tasks, even compared with simple baselines like chain-of-thought.
    Even when scaling model capabilities via more parameters or generation iterations, LCEs do not consistently boost existing performance. 
    Across all datasets and models, the performance gains of LCEs are negatively correlated ($r=-0.5, p\le 0.05$) with the diversity of the failure modes of reasoning trajectories.
    % LCEs are statistically more useful ($p\le 0.05$) when the failure modes of reasoning in a dataset are less diverse.
    Investigating how LCEs work, we find that (1) the improvement in instruction adherence can explain some success of LCEs; 
    (2) one major bottleneck is whether LLMs can extract effective experiences out of natively diverse failures. 
    Through this study, we uncover the limited capabilities of LCEs in abstracting principled contexts from diverse failure modes and understand how LCEs work, providing guidelines for future work. 
}

\maketitle

% ------------- input files -------
%!TEX root = ../main.tex
\section{Introduction}
\label{sec:intro}

Recent advances in LLMs have demonstrated strong generalization ability across a wide range of real-world applications.
In parallel with these developments, one line of work adapts LLMs to downstream tasks through training-based methods such as Reinforcement Learning from Human Feedback (RLHF)~\citep{RLHF}, Direct Preference Optimization (DPO)~\citep{DPO}, and Group Relative Policy Optimization (GRPO)~\citep{GRPO}, which update the model parameters to align with task-specific objectives. However, these training-based methods often demand large amounts of preference data~\citep{RLHF} or hundreds of thousands of rollouts~\citep{GRPO}. 
The inefficiency of these methods motivated a complementary line of work that seeks to adapt LLMs in a training-free manner by engineering the prompt or context of LLMs rather than altering their intrinsic capabilities.

% Early efforts
Early efforts in this direction mainly rely on human heuristics to engineer the instructions~\citep{GEPA,TFGRPO,DSPy,MIPROv2,TextGrad,ACE} that elicit LLM reasoning capabilities~\citep{SC,CoT,ToT} or execute a better workflow~\citep{ADAS,Combee,Reflexion,metaTextGrad,ReAct,Aflow,AdaptFlow}. In-Context Learning (ICL)~\citep{GPT3} enables few-shot learning via in-context examples. Chain-of-Thought (CoT) prompting and its variants~\citep{CoT,ToT} instruct LLMs to generate intermediate reasoning steps.
% which significantly improved performance. 
Later, agent frameworks~\citep{ReAct} rely on the reasoning to take actions, allowing LLMs to interact with external tools and environments.
However, such methods are designed to accomplish tasks in a single attempt but cannot learn from accumulated human trials recorded in offline datasets. 
For example, CoT may avoid a mistake in one attempt through reasoning, but the same error may recur in subsequent trials.

Given a set of training examples, \textbf{LLM as Context Engineer (LCE)}, or automatic context engineering, was proposed to learn context from data without parameter training~\citep{TFGRPO, ADAS, A-MEM, yang2024large, ACE}.
% \junyuan{Here you should cite many papers of LLM auto context engineering. The most important ones. At least 5...} \jr{done}.
The main principle is to cast the LLM itself as a context engineer that operates on textual space rather than parameter space~\citep{yang2024large}.
Multi-prompt Instruction Proposal Optimizer v2 (MIPROv2)~\citep{MIPROv2} and Genetic-Pareto (GEPA)~\citep{GEPA} propose effective sampling strategies for searching over candidate prompts.
Training-Free Group Relative Policy Optimization (TF-GRPO)~\citep{TFGRPO} extracts semantic advantages by comparing agent rollouts, while Agentic Context Engineering (ACE)~\citep{ACE} curates an evolving context and uses metadata to track the contributions of each insight.

\begin{figure*}[t]
    \centering
    % \vspace{-0.1in}
    \includegraphics[width=\textwidth]{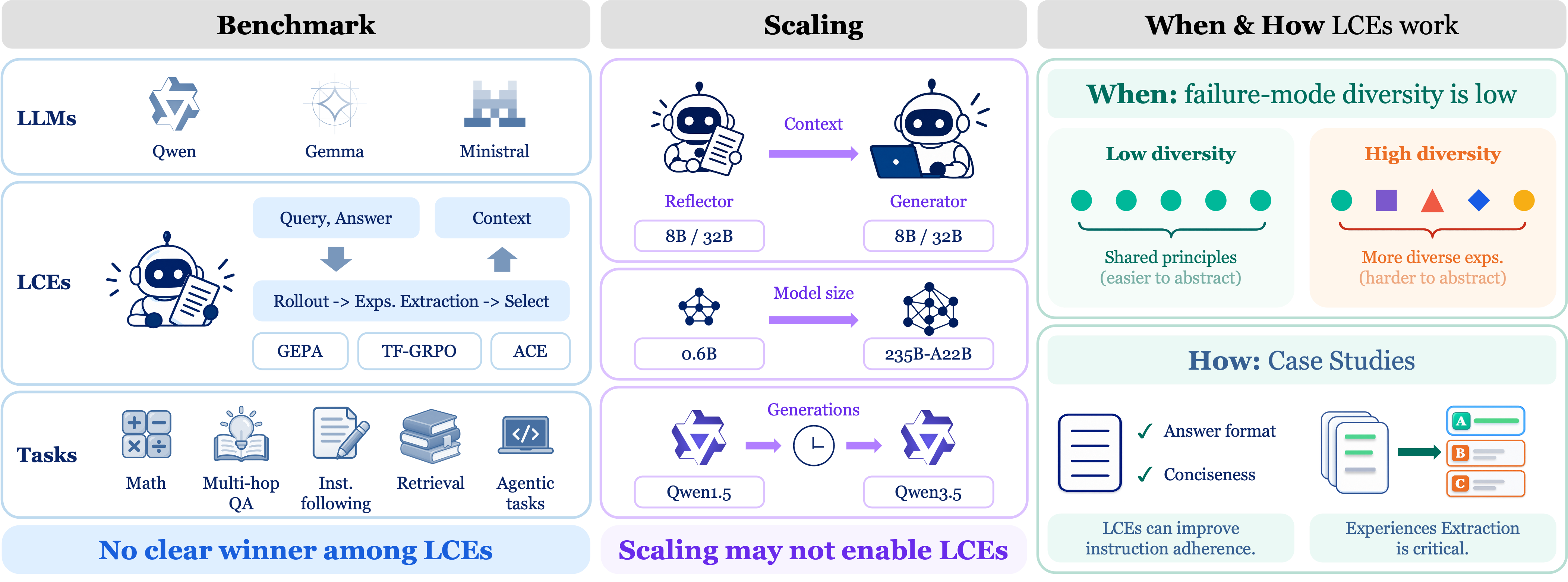}
    \vspace{-0.22in}
    \caption{Overview of our work. We investigate whether LCEs can engineer effective context and further examine the mechanisms underlying their successes and failures.
    } 
    \vspace{-0.18in}
    \label{fig:overview}
\end{figure*}

% \junyuan{what is the gap? why the gap is important.} \jr{done}
Despite their success, these methods are typically evaluated using different LLMs and distinct benchmarks, leaving their relative advantages and broader applicability unclear. 
For example, ACE~\citep{ACE} reports improvements with DeepSeek-V3.1-671B~\citep{DeepSeek-V3} on agentic ~\citep{AppWorld} and financial tasks~\citep{FiNER, Formula}, whereas GEPA~\citep{GEPA} reports improvements with Qwen3-8B~\citep{Qwen3} and GPT-4.1-mini~\citep{openai2025gpt41} on math~\citep{MATH} and multi-hop reasoning~\citep{HotpotQA, HoVer}. 
These inconsistencies in evaluation protocols motivate us to conduct a comprehensive benchmark of LCEs across LLM families and diverse tasks under comparable evaluation settings.
As summarized in~\cref{fig:overview}, our major contributions in this paper are as follows. (1) We conduct a comprehensive comparison between LCE methods in a unified benchmark, using various LLMs, LCE methods, and task types. (2) We also evaluate the effect of scaling model parameter size and model generations. (3) We conduct new ablation studies and case studies to show when and how LCEs can actually work.
% Our results show that no single method can consistently outperform even CoT prompting, across datasets or models.
% Scaling experiments then indicate that larger models can amplify existing gains but cannot recover from failures. Meanwhile, scaling to more parameters and newer LLM generations are insufficient to improve LCE effectiveness.
% % These results lead us to ask when and how LCEs can engineer the effective context. 
% % Through detailed analysis and extensive ablation studies, we observe that LCEs are more effective when facing more diverse failure modes and then provide empirical support for this observation.
% % Further decomposition by factor shows that variance in failure-mode diversity primarily comes from dataset differences. 
% These results lead us to ask when and how LCEs can engineer effective context. Through extensive investigation, we find that LCEs tend to be more effective when facing less diverse failure modes. 
% We further observe that LCEs struggle to reduce experience diversity, helping explain their limited effectiveness on tasks with more diverse failure modes. Our ablations also show that experience extraction can discard useful knowledge from demonstrations. Together, these findings highlight a key bottleneck: the limited ability of current LCEs to distill diverse failures into reusable principles for effective context.
Our key findings are as follows:
% \jr{Is the current intro (especially the last paragraph) and the key findings OK?}
\begin{itemize}[leftmargin=0.2in]
    % \item \textbf{SLM can} engineer contexts for the retrieval tasks \textbf{but} not on reasoning-intensive tasks.
    % \item \textbf{SLM can} engineer contexts to introduce new knowledge, \textbf{but} the introduced knowledge may also be harmful. Our experiments suggest that different Context Engineers can improve the performance for the retrieval tasks, while not in more reasoning-intensive tasks. 
    \item \textbf{LLMs can} engineer contexts via task-specific training, but no single method can achieve the best performance in all datasets at test time.
    \item \textbf{Scaling} helps but cannot reverse existing failures of LCEs. Even scaling to newer generations and more parameters remains insufficient to make LCEs more effective across different settings.
    \item \textbf{When\&How:} Through aggregate analysis, we find that LCEs work better when facing less diverse failure modes of reasoning. Case studies show that the ability to abstract principled contexts from heterogeneous failures is the major bottleneck and that the success of some LCE methods is attributed to the improvement in adherence to (format) instruction rather than reasoning.
    % \item \textbf{When and how} do LCEs help? Through careful analysis and extensive ablation studies, we find that LCEs succeed when trajectories contain less diverse failure modes, and we further identify the main bottleneck of current LCEs as their limited capacity to abstract principled contexts from heterogeneous failures.
\end{itemize}
% Composing these insights, we suggest to (1) use a relatively larger LM to engineer the context and inference with SLM and (2) use RAG at test time to dynamically reduce the context. 

% Specifically, on MiddleMatch, we show that combining (1) and (2) consistently improves TF-GRPO's performance. 
% \junyuan{I remember I said that if you don't have the results, do not claim this. Where is your results of combining 1 and 2??? If I am the reviewer, I will recommend desk rejection on this.} \jr{will be added after providing the positive results} \jr{Running RAG k=1/3/5/10 now}

% \junyuan{As our claim is on SLM (note the `S'), we need to use term like LM (without Large) Context Engineer instead of LLM Context Engineer. Some acronym like LCE could be used.} \jr{I see}

%!TEX root = ../main.tex
\section{Related Work}
\label{sec:related}

\textbf{AI Agents.}
AI Agent frameworks can be viewed as test-time scaling methods that allocate additional inference-time computation to organize the operation of LLMs during task execution, without modifying the model parameters.
CoT prompting~\citep{CoT} is a widely adopted starting point for eliciting the reasoning capabilities of LLMs by generating intermediate steps before the final answer, achieving substantial improvements across diverse downstream applications.
Building on CoT, self-consistency~\citep{SC} samples multiple reasoning chains from the same prompt and selects the final answer by majority voting to improve reasoning reliability.
Beyond purely internal reasoning, ReAct~\citep{ReAct} interleaves reasoning traces with task-specific actions such as tool calls, allowing the model to ground its reasoning in external observations rather than relying solely on parametric knowledge.
Another research line extends reasoning into more structured trajectories. Tree-of-Thoughts~\citep{ToT} frames reasoning as a search over a tree of intermediate thoughts, allowing the model to explore multiple reasoning paths.
Graph-of-Thoughts~\citep{GoT} further extends this paradigm by representing intermediate thoughts as nodes in a reasoning graph, enabling more flexible aggregation, recombination, and refinement of partial solutions.
Despite this progress, these frameworks rely on predefined workflows and offer limited flexibility in adapting to specific tasks.
Our study therefore focuses on analyzing methods that automatically engineer contexts in a training-free manner.

\noindent\textbf{LLM Context Engineers (LCEs).}
% \junyuan{Some important early works are missing, e.g., \citep{yang2023large}, \citep{zhou2022large,sordoni2023joint} } \jr{done}
Similar to training-based methods, LCEs follow the same optimization loop but rely on textual feedback and updates to the context text rather than gradient-based parameter modifications. Automatic Prompt Engineer~\citep{zhou2022large,hong2024dp,yang2026lad} pioneers this paradigm by searching over LLM-generated instruction candidates to maximize task accuracy. Optimization by Prompting~\citep{yang2024large} casts the LLM itself as an iterative optimizer that refines prompts based on past trajectory. Deep Language Networks~\citep{sordoni2023joint} extend this idea to stacked LLMs by jointly optimizing prompts via variational inference. Reflexion~\citep{Reflexion} introduces a self-reflection feedback loop where the model verbally reflects on task outcomes and conditions subsequent attempts on these self-generated critiques.
% Building on these advancements, subsequent methods move beyond instance-specific reflection toward optimizing a reusable system prompt that encodes a general strategy for the target task.
% DSPy~\citep{DSPy} provides a unified toolkit inspired by PyTorch abstractions, enabling subsequent methods such as GEPA~\citep{GEPA} and MIPROv2~\citep{MIPROv2} to optimize few-shot demonstrations \junyuan{does GEPA optimize few-shot examples??} and iteratively update the system prompt. In parallel\junyuan{they are not in prallel at all... Please at read their publication time, before you cite.}, TextGrad~\citep{TextGrad} extends the computational graph abstraction from gradient back-propagation to natural language, propagating textual gradients via a textual chain rule to optimize prompts in a structured pipeline.
 % \junyuan{does GEPA optimize few-shot examples??} \junyuan{they are not in prallel at all... Please at read their publication time, before you cite.}
Building on these advancements, subsequent methods move beyond instance-specific reflection toward optimizing a reusable system prompt that encodes a general strategy for the target task. 
DSPy~\citep{DSPy} provides a unified toolkit inspired by PyTorch abstractions for declaratively programming LLM pipelines. 
% Within this framework, MIPROv2~\citep{MIPROv2} jointly optimizes instructions and few-shot demonstrations through Bayesian optimization, while the more recent GEPA~\citep{GEPA} adopts reflective prompt evolution with Genetic-Pareto search, optimizing instructions through natural-language reflection on rollouts.
Within this framework, MIPROv2~\citep{MIPROv2} and GEPA~\citep{GEPA} propose different sampling strategies to optimize the instructions.
Inspired by backpropagation in neural networks, TextGrad~\citep{TextGrad} uses natural language to propagate textual gradients. 
Recently, another line of work maintains an evolving external context, such as abstract experiences or reusable skills~\citep{FLEX, AutoGEO, Combee}. TF-GRPO~\citep{TFGRPO} and ReasoningBank~\citep{ReasoningBank} contrast successful and failed task executions to extract reusable insights. ACE~\citep{ACE} curates an evolving playbook through iterative bullet-level updates.
Despite the success in the reported benchmarks, the inconsistent evaluation makes the actual effectiveness of LCEs unclear. Thus, this work aims to uncover and understand the limitations of LCEs.

\section{Preliminaries}
\label{sec:prelim}

When prompting LLM to answer a question, the input contains not only a query but also a context. The \textbf{context} could be an instruction or essential information, e.g., definitions of terminology, for answering the query.
Distinct from manual context engineering, \textbf{LCEs} generate context automatically by iteratively updating the context based on information gained from data. 
In this paper, we conceptualize the paradigm as a way to learn context $\pi$ (a sequence of tokens) from a dataset $D=\{(x,y)\}$ where $(x,y)$ is a pair of query and response to the LLM. We are interested in investigating whether LLMs can learn a generalizable $\pi$ (prompt, experiences, or metadata) from $D$.
% Input dataset $D=\{(x,y)\}$, output context $\pi$ which could be prompt, experience pool or metadata.
Context engineering can be formulated as
$\pi^* = \argmax_{\pi} V(\pi; D)$,
% Context engineer can be formulated as
% % LCEs can be formulated as
% \begin{align*}
%     \pi^* = \argmax _\pi V(\pi; D)
% \end{align*}
where $V$ quantifies the performance of $\pi$.
A general context engineering method can be formulated as
\begin{align}
    \pi=\phi_V(\{E(x,t,y) | t\sim \text{LLM}(x,\pi^-), (x,y)\sim D\} | \pi^- \sim \Pi). \label{eq:po}
\end{align}

The formulation includes three steps that transform samples into an informative context $\pi$. \textbf{Step R (Rollout):} a trajectory $t$ is unrolled to answer the query $x$ based on a sampled context $\pi^{-}$. \textbf{Step E (Extraction):} $E(\cdot)$ processes the triplet $(x,t,y)$ to generate an intermediate artifact, e.g., experience or feedback. \textbf{Step $\phi$ (Consolidation):} $\phi_V(\cdot)$ leverages the evaluation $V$ to transform the intermediate artifact to produce the final context by consolidating the most important information.

Existing context engineering methods instantiate the~\cref{eq:po} in different ways.
Below, we review four representative methods with open-source implementations.
% \\
% \noindent\textbf{MIPROv2}~\citep{MIPROv2} jointly optimizes instructions and few-shot demonstrations through Bayesian sampling over candidate prompts. Intuitively, MIPROv2 generates a batch of candidate prompts and selects the prompts based on their validation performance per iteration. 
% In \cref{eq:po}, $\Pi$ is the distribution of prompts conditioned on samples. $E$ is the identity function. $t$ is simply the predicted $y$. $\phi_V$ selects the prompt with the best validation performance.
% In this work, we use the official implementation in the DSPy package~\citep{DSPy} with \texttt{auto="heavy"} to maximize performance.
\\
\noindent\textbf{TextGrad}~\citep{TextGrad} propagates textual ``gradients'' (reflection) through a structured pipeline to update prompts iteratively. The method integrates agentic reflection~\citep{yang2024large} in a back-propagation form, enabling optimization of hierarchical context.
For TextGrad, $\Pi$ is the delta distribution on the prompt from the last iteration. $E$ generates feedback on the prompt and $\phi_V$ updates the prompt based on the feedback.
%according to the feedback.
\\
\noindent\textbf{GEPA}~\citep{GEPA} realizes a sample-efficient prompt optimizer for compound AI systems. GEPA samples execution trajectories and uses an LLM to reflect on them in natural language to diagnose failures. To avoid local optima, GEPA samples new prompt candidates from a Pareto frontier of high-performing generated prompts in training.
We can formulate GEPA in \cref{eq:po}, where $\Pi$ is a distribution characterized by the Pareto frontier in the search tree, $E$ conducts reflective prompt mutation based on the feedback on $t$, and $\phi_V$ updates the prompt.
\\
\noindent\textbf{TF-GRPO}~\citep{TFGRPO} instantiates the agentic reinforcement learning. 
In contrast to GEPA's iterative rewriting of the system prompt, TF-GRPO maintains an external experience pool, instantiating the policy as a frozen LLM paired with a variable experiential context. 
Mirroring vanilla GRPO~\citep{GRPO}, TF-GRPO generates grouped rollouts in training but replaces the numerical group-relative advantage with semantic reflection on trajectory differences. 
TF-GRPO instantiates \cref{eq:po} as follows: $E$ extracts the experiences via reflecting on $t$, and $\phi$ consolidates experiences into the pool as context.
\\
\noindent\textbf{ACE}~\citep{ACE} focuses on augmenting external information in context -- a structured playbook containing items, each carrying \emph{metadata} (a unique identifier and counters tracking helpfulness) and \emph{content} (a short reusable strategy or failure mode). ACE applies incremental delta updates iteratively and only consolidates the playbook by merging semantically similar entries via embedding-based de-duplication, without actively deleting past bullets. 
ACE is similar to TF-GRPO in \cref{eq:po}, except that the context $\pi$ is composed of metadata and more structured experiences.

\begin{wraptable}{r}{0.49\linewidth}
  \centering
  \vspace{-0.15in}
  \caption{\textbf{Comparison of evaluated context engineering methods.}
  \emph{Sys} denotes the system prompt; \emph{Exp} denotes experience
  (e.g., a pool of distilled insights or a playbook injected into context).}
  \vspace{-0.1in}
  \label{tab:method-compare}

  \footnotesize
  \setlength{\tabcolsep}{2pt}
  \renewcommand{\arraystretch}{1.05}

  \begin{tabular}{@{}
    >{\raggedright\arraybackslash}p{0.20\linewidth}
    >{\centering\arraybackslash}p{0.3\linewidth}
    >{\raggedright\arraybackslash}p{\dimexpr0.5\linewidth-4\tabcolsep\relax}
  @{}}
    \toprule
    \textbf{Method}
    & \textbf{Context $\pi$}
    & \textbf{Main Strategy} \\
    \midrule
    TextGrad
    & Sys
    & Textual gradient backpropagation \\[2pt]
    GEPA
    & Sys
    & Textual reflection + Mutation sampling \\[2pt]
    TF-GRPO
    & Exp
    & Group-relative textual advantage \\[2pt]
    ACE
    & Exp + metadata
    & Textual reflection \\
    \bottomrule
  \end{tabular}
  \vspace{-0.15in}
\end{wraptable}

As summarized in \cref{tab:method-compare}, these methods cover a wide spectrum of principles. 
% conditioned prompt sampling (MIPROv2~\citep{MIPROv2}), LLM-driven textual reflection on prompt updates (TextGrad~\citep{TextGrad}), GEPA~\citep{GEPA}, agentic reinforcement learning (TF-GRPO)~\citep{TFGRPO}, and agentic context generation (ACE~\citep{ACE}).
In terms of the form of $\pi$, TextGrad, and GEPA were designed for engineering the system prompts, while TF-GRPO and ACE were designed for distilling experiences that augment the context at test time.

In our study, we also include two simple baselines that do not require iterative engineering.
(1) Chain-of-Thought \textbf{(CoT)}~\citep{CoT} uses the prompt \emph{"Solving the problem step by step, \{Question or Other Input Context\}"} to elicit reasoning before the final prediction, while for AppWorld, we use the initial prompt from the ACE implementation to specify the basic constraints and rules.
% \noindent\textbf{Random-ICL}~\citep{GPT3} uses successful few-shot demonstrations to illustrate reasoning strategies or critical actions for task completion. We first solve each training sample five times to build a pool of reasoning trajectories. For each test sample, we then randomly select three successful trajectories from this pool as in-context demonstrations. 
(2) In-Context Learning \textbf{(ICL)}~\citep{GPT3} guides LLMs using successful rollouts. Different from \citet{GPT3}, we enrich the context with reasoning demonstrations to illustrate how to complete a task step-by-step. We first generate five trajectories for each training sample to construct a demonstration pool. For each test sample, we then randomly select three successful trajectories from this pool as in-context demonstrations.

\section{Can LLMs Engineer Context?}
\label{sec:benchmark-results}

We reviewed the evaluation protocols used in different publications of the representative methods and observed minimal overlap across them (see details in~\cref{tab:dataset_coverage}).
The disjoint datasets leave the general effectiveness of context engineering unclear.
To understand the effectiveness of LCEs, we establish a unified benchmark across different LLM families in 8B size. We include diverse tasks: 
mathematical reasoning (MATH500,~\cite{MATH500}), multi-Hop reasoning (HotpotQA,~\cite{HotpotQA}), instructions following (IFBench,~\cite{IFbench}), retrieval (MiddleMatch,~\cite{MiddleMatch}) and Agentic Tasks (AppWorld, ~\cite{AppWorld}).
Each experiment is repeated 5 times with 5 fixed seeds.
Additional experimental setup is provided in~\cref{appendix:setup}.

% To investigate the behaviors exhibited by these methods under a fair and rigorous comparison, we first establish a unified benchmark across different LLM families in 8B size and diverse tasks, from mathematical reasoning to agentic application, to investigate the effectiveness of the LCEs. 
% We select various datasets with different reasoning patterns

\begin{figure*}[t]
    \centering    
    \includegraphics[width=0.95\textwidth]{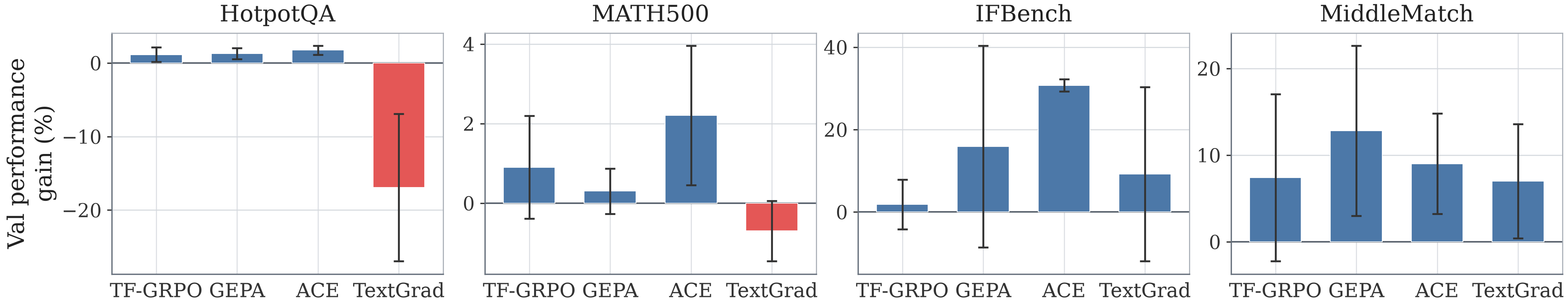}
    \vspace{-0.1in}
    \caption{Validation performance gains after context engineering in 4 datasets. 
    }
    \label{fig:benchmark_po}
    \vspace{-0.1in}
\end{figure*}

\begin{table*}[t]
    % \vspace{-0.05in}
    \caption{Performance of LCEs across LLM families. No method consistently improves the context. \textbf{Bold} and $(\cdot)$ denote the best performance and performance changes relative to the initial, respectively. In the $\Delta_{\text{best}}$ column, $*$ denotes a significant difference ($p \leq 0.05$); significant improvements and degradations are shaded blue and red, respectively.
    }
    \vspace{-0.05in}
    \label{tab:benchmark_all_models}
    \centering
    \footnotesize
    \setlength{\tabcolsep}{2pt}
    \resizebox{\textwidth}{!}{%
    \begin{tabular}{ll|cc|ccc|c}
        \toprule
        \textbf{Model} & \textbf{Dataset}
        & \textbf{CoT} & \textbf{ICL}
        & \textbf{TF-GRPO} & \textbf{GEPA} & \textbf{ACE}
        & $\boldsymbol{\Delta_{\text{best}}}$ \\
        \midrule

        \multirow{5}{*}{Qwen3}
        & MiddleMatch
        & $31.27 \pm 2.10$
        & $\mathbf{46.07 \pm 4.80}$
        & $37.73 \pm 8.57$ {\scriptsize \textcolor{black!70}{(+6.47)}}
        & $42.33 \pm 3.80$ {\scriptsize \textcolor{black!70}{(+12.53)}}
        & $25.20 \pm 3.96$ {\scriptsize \textcolor{black!70}{(+4.73)}}
        & $-3.74$ \\

        & IFBench
        & $32.35 \pm 0.82$
        & $30.54 \pm 0.94$
        & $32.21 \pm 1.58$ {\scriptsize \textcolor{black!70}{(-0.14)}}
        & $\mathbf{33.30 \pm 3.64}$ {\scriptsize \textcolor{black!70}{(-0.37)}}
        & $32.17 \pm 1.28$ {\scriptsize \textcolor{black!70}{(+0.64)}}
        & $+0.95$ \\

        & MATH500
        & $\mathbf{83.20 \pm 2.51}$
        & $79.80 \pm 1.04$
        & $82.10 \pm 2.43$ {\scriptsize \textcolor{black!70}{(-1.10)}}
        & $80.60 \pm 1.43$ {\scriptsize \textcolor{black!70}{(-0.20)}}
        & $63.80 \pm 2.93$ {\scriptsize \textcolor{black!70}{(-7.30)}}
        & $-1.10$ \\

        & HotpotQA
        & $\mathbf{38.00 \pm 0.75}$
        & $37.60 \pm 0.83$
        & $\mathbf{38.00 \pm 0.97}$ {\scriptsize \textcolor{black!70}{(0.00)}}
        & $32.80 \pm 0.99$ {\scriptsize \textcolor{black!70}{(-1.54)}}
        & $33.67 \pm 0.97$ {\scriptsize \textcolor{black!70}{(-3.26)}}
        & $0.00$ \\

        & AppWorld
        & $21.55 \pm 1.95$
        & $23.10 \pm 3.89$
        & $\mathbf{23.69 \pm 2.08}$ {\scriptsize \textcolor{black!70}{(+2.14)}}
        & $23.33 \pm 1.54$ {\scriptsize \textcolor{black!70}{(+0.48)}}
        & $18.93 \pm 1.29$ {\scriptsize \textcolor{black!70}{(-3.45)}}
        & $+0.59$ \\
        \midrule

        \multirow{5}{*}{Gemma4}
        & MiddleMatch
        & $79.33 \pm 1.93$
        & $80.47 \pm 2.32$
        & $90.13 \pm 3.43$ {\scriptsize \textcolor{black!70}{(+10.80)}}
        & $\mathbf{93.33 \pm 2.91}$ {\scriptsize \textcolor{black!70}{(+7.40)}}
        & $87.47 \pm 6.16$ {\scriptsize \textcolor{black!70}{(+0.73)}}
        & \cellcolor{blue!50}$+12.86^{*}$ \\

        & IFBench
        & $39.76 \pm 2.22$
        & $34.63 \pm 2.17$
        & $41.63 \pm 1.21$ {\scriptsize \textcolor{black!70}{(+1.87)}}
        & $\mathbf{46.70 \pm 4.54}$ {\scriptsize \textcolor{black!70}{(-1.22)}}
        & $42.09 \pm 1.43$ {\scriptsize \textcolor{black!70}{(+1.13)}}
        & \cellcolor{blue!29}$+6.94^{*}$ \\

        & MATH500
        & $67.70 \pm 1.04$
        & $\mathbf{77.90 \pm 3.54}$
        & $73.80 \pm 4.12$ {\scriptsize \textcolor{black!70}{(+6.10)}}
        & $71.20 \pm 1.92$ {\scriptsize \textcolor{black!70}{(+3.50)}}
        & $69.40 \pm 1.14$ {\scriptsize \textcolor{black!70}{(+1.60)}}
        & $-4.10$ \\

        & HotpotQA
        & $\mathbf{33.80 \pm 0.80}$
        & $32.60 \pm 3.27$
        & $33.67 \pm 1.29$ {\scriptsize \textcolor{black!70}{(-0.13)}}
        & $29.47 \pm 1.45$ {\scriptsize \textcolor{black!70}{(+0.73)}}
        & $31.67 \pm 0.78$ {\scriptsize \textcolor{black!70}{(-0.20)}}
        & $-0.13$ \\

        & AppWorld
        & $25.24 \pm 2.25$
        & $26.67 \pm 2.50$
        & $\mathbf{29.05 \pm 3.13}$ {\scriptsize \textcolor{black!70}{(+3.81)}}
        & $28.81 \pm 0.68$ {\scriptsize \textcolor{black!70}{(+1.67)}}
        & $26.07 \pm 4.40$ {\scriptsize \textcolor{black!70}{(-0.36)}}
        & $+2.38$ \\
        \midrule

        \multirow{5}{*}{Min3}
        & MiddleMatch
        & $62.00 \pm 3.44$
        & $49.53 \pm 3.91$
        & $\mathbf{71.40 \pm 6.46}$ {\scriptsize \textcolor{black!70}{(+9.40)}}
        & $59.60 \pm 4.64$ {\scriptsize \textcolor{black!70}{(+22.20)}}
        & $38.93 \pm 8.69$ {\scriptsize \textcolor{black!70}{(-1.07)}}
        & \cellcolor{blue!38}$+9.40^{*}$ \\

        & IFBench
        & $28.50 \pm 2.19$
        & $26.46 \pm 1.87$
        & $28.57 \pm 1.88$ {\scriptsize \textcolor{black!70}{(+0.07)}}
        & $\mathbf{36.70 \pm 1.84}$ {\scriptsize \textcolor{black!70}{(+0.82)}}
        & $31.21 \pm 1.90$ {\scriptsize \textcolor{black!70}{(+0.27)}}
        & \cellcolor{blue!34}$+8.20^{*}$ \\

        & MATH500
        & $87.50 \pm 1.06$
        & $68.10 \pm 2.41$
        & $\mathbf{88.30 \pm 1.15}$ {\scriptsize \textcolor{black!70}{(+0.80)}}
        & $85.10 \pm 1.43$ {\scriptsize \textcolor{black!70}{(-1.30)}}
        & $80.10 \pm 3.90$ {\scriptsize \textcolor{black!70}{(-3.00)}}
        & $+0.80$ \\

        & HotpotQA
        & $23.07 \pm 1.26$
        & $27.07 \pm 0.95$
        & $24.13 \pm 1.91$ {\scriptsize \textcolor{black!70}{(+1.07)}}
        & $31.00 \pm 1.60$ {\scriptsize \textcolor{black!70}{(+2.80)}}
        & $\mathbf{31.27 \pm 1.38}$ {\scriptsize \textcolor{black!70}{(+3.13)}}
        & \cellcolor{blue!20}$+4.20^{*}$ \\

        & AppWorld
        & $26.43 \pm 1.96$
        & $28.45 \pm 2.12$
        & $27.14 \pm 2.75$ {\scriptsize \textcolor{black!70}{(+0.71)}}
        & $24.29 \pm 7.52$ {\scriptsize \textcolor{black!70}{(-1.90)}}
        & $\mathbf{30.60 \pm 1.08}$ {\scriptsize \textcolor{black!70}{(+4.40)}}
        & $+2.15$ \\

        \bottomrule
    \end{tabular}
    }
    \vspace{-0.2in}
\end{table*}

\par\noindent
\textbf{LCEs generally improve performance in training.} In \cref{fig:benchmark_po}, we compare how different LCEs can improve the context \emph{before} and \emph{after} engineering, namely, during training. We report their validation performance gains and use the same Qwen3-8B (Qwen3) model~\citep{Qwen3}, where performance gain is defined as the performance score difference between after and before LCE. The validation set was used to select engineered context during training.
Our results show that TF-GRPO, GEPA, and ACE consistently improve validation performance across the four datasets. On MiddleMatch and IFBench, all LCEs improve validation performance, with maximum gains exceeding 20 points for ACE on IFBench and 10 points for GEPA. On MATH500 and HotpotQA, TF-GRPO, GEPA, and ACE achieve only marginal validation gains, while TextGrad fails to produce effective contexts and even leads to a substantial performance decline of more than 15 points on HotpotQA.

% TF-GRPO, GEPA, and ACE improve validation accuracy across all four datasets, whereas TextGrad exhibits substantial declines on HotpotQA and MATH500.
\par\noindent
\textbf{No free lunch for LCEs at test time.}
Beyond validation, we compare these methods at test time, where we only use three methods (TF-GRPO, GEPA, ACE) that stably improve validation performance. 
To evaluate LCEs in multi-step agentic settings, we additionally include AppWorld, a benchmark requiring LLMs to complete long-horizon tasks to further assess whether LCEs remain effective when task completion requires sustained interaction rather than a single response.
We include three LLM families in our evaluation: Qwen3-8B (Qwen3), Gemma-4-E4B-it (Gemma4)~\citep{team2026gemma}, and Ministral-3-8B-Instruct-2512-BF16 (Min3)~\citep{liu2026ministral}.
\cref{tab:benchmark_all_models} reports the performance on different datasets across three LLM families and the performance changes relative to each LCE's initial, with CoT and ICL included as reference baselines.
\\
% \jr{(1)mean>1 (2)significant (3)below than initial} 
\emph{(1) LCEs provide limited advantages over simple baselines.} 
    Although these LCEs can improve validation performance, these gains do not consistently transfer to the test set, where LCEs can even underperform simple baselines, or the gains cannot be attributed to the algorithm itself. 
    To assess LCEs' effectiveness on the test set, we first examine the absolute performance differences between the best-performing LCE and the best baseline, reported in the $\Delta_{\text{best}}$ column of~\cref{tab:benchmark_all_models}, and find that only 7 of the 15 settings show observable improvements ($\Delta_{\text{best}}>1$). We further examine these gains using statistical significance tests, which indicate that only 5 settings show significant improvements ($p<0.05$). Even among these settings, we observe that some \emph{gains stem from the initial context design} rather than from engineering a better context during training, through comparisons between the final and initial performance of the best-performing LCEs. For example, GEPA on IFBench with Gemma4 surpasses CoT by 6.94 points but remains 1.22 points below its initial performance.
    % Although these LCEs can improve the validation performance, these gains do not consistently transfer to the test set and even cause degradation relative to simple baselines. For instance, all LCEs fail to match CoT prompting on MATH500 with Qwen3. 
    % (b) We further investigate if LCEs actually bring new knowledge or just change the chance of success. Additional evaluations using Pass@K (success in at least one of K repeated samples) and \text{Pass\textasciicircum K} (success in all K repeated samples) in Appendix~\ref{appendix:passk} further suggest that LCEs may introduce harmful knowledge that impairs reasoning on difficult problems in ways that repeated sampling cannot reverse and undermines the model’s ability to consistently produce correct answers on simpler problems.
    % \jr{leo:confused about Pass@K and \text{Pass\textasciicircum K}. Should we delete Pass@K?} \jr{I changed the (explanation) after metrics to repeated samples. }
    Meanwhile, we notice there are some wins of LCEs over CoT or ICL for specific models on HotpotQA and MATH500. But the wins occur only when a model’s baseline performance is lower than that of other models on the same task. 
    For example, on HotpotQA, Min3 has the lowest CoT and ICL scores among the three models (23.07 and 27.07, respectively), while ACE raises its score by 3.13 compared to its initial. Yet, LCEs yield no performance gains for the other LLMs. Thus, we conjecture that such gains are more likely when the model’s capabilities are not fully exploited.
\\
\emph{(2) No clear winner among LCEs across datasets or models.}
    For example, TF-GRPO performs best on AppWorld with Qwen3, while ACE leads on Min3. 
    % These results suggest that the randomness of the methods dominates the results.
    % \jr{leo finds this paragraph misleading, because different methods use different initial prompts, so some methods improve performance but less than other LCEs. Would “No method consistently improves performance” be clearer? However, this point is already covered under \textbf{Limited advantages over simple baselines}. Should we combine these two points?}
% \\
% \emph{(3)} When comparing performance gains across LLM families, we find that the effectiveness of LCEs cannot generalize.
We also observe that LCEs substantially improve Gemma4 on MATH500 but provide limited performance gains for Qwen3 and Min3.
On HotpotQA, LCEs improve Min3 more than the other models.
The limitation suggests that there is no free lunch for LCEs -- different models/datasets may need different LCE algorithms.

The limited generalizability of LCE methods encourages us to further investigate the working mechanism in the later sections.

\par
\begin{figure*}[t]
    \centering
    % \vspace{-0.05in}
    \includegraphics[width=\textwidth]{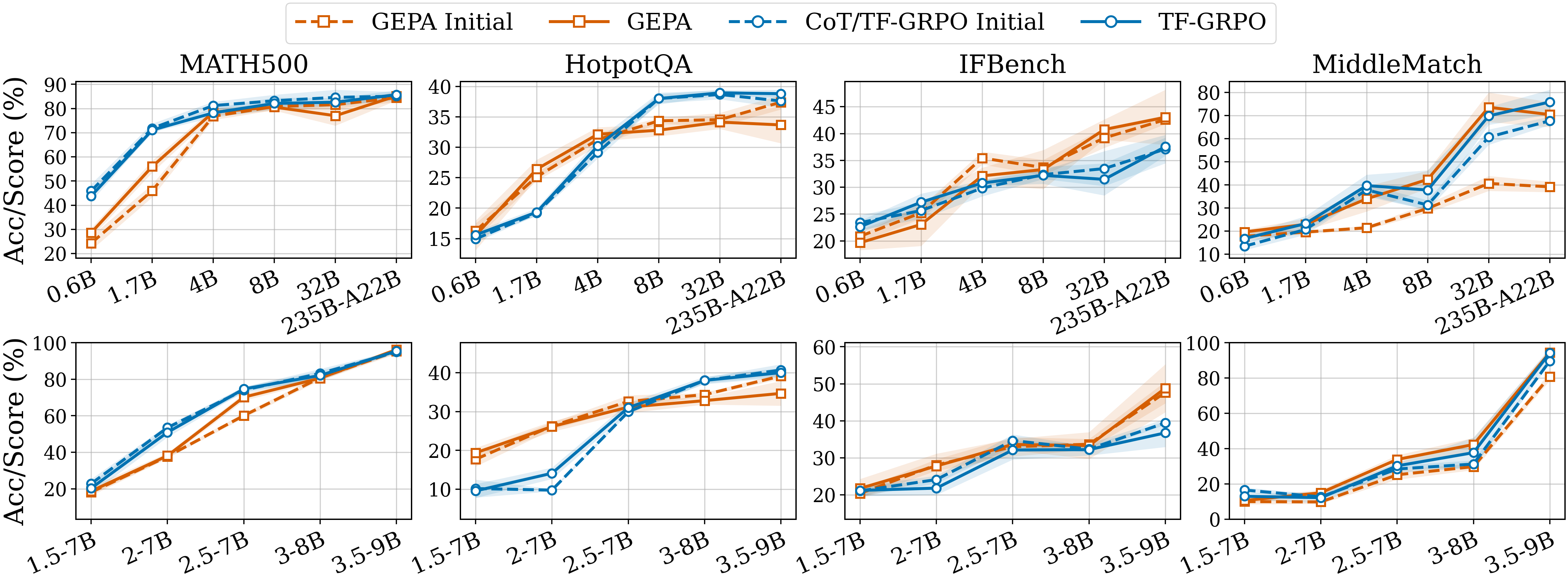}
    \vspace{-0.2in}
    \caption{Scaling experiments. \textbf{Top:} Parameter-size scaling of self-reflection from 0.6B to 235B-A22B (Qwen3) across different datasets. \textbf{Bottom:} Comparison of model generations from Qwen1.5 to Qwen3.5.
    CoT and GEPA initial are the initials for TF-GRPO and GEPA, respectively.
    } 
    \vspace{-0.2in}
    \label{fig:gepa_scaling}
\end{figure*}

\noindent
\textbf{LCE with Scaling.}
% \jr{Here may also need a paragraph to describe what we done in this section}
\begin{figure*}[t]
    \centering
    \includegraphics[width=\textwidth]{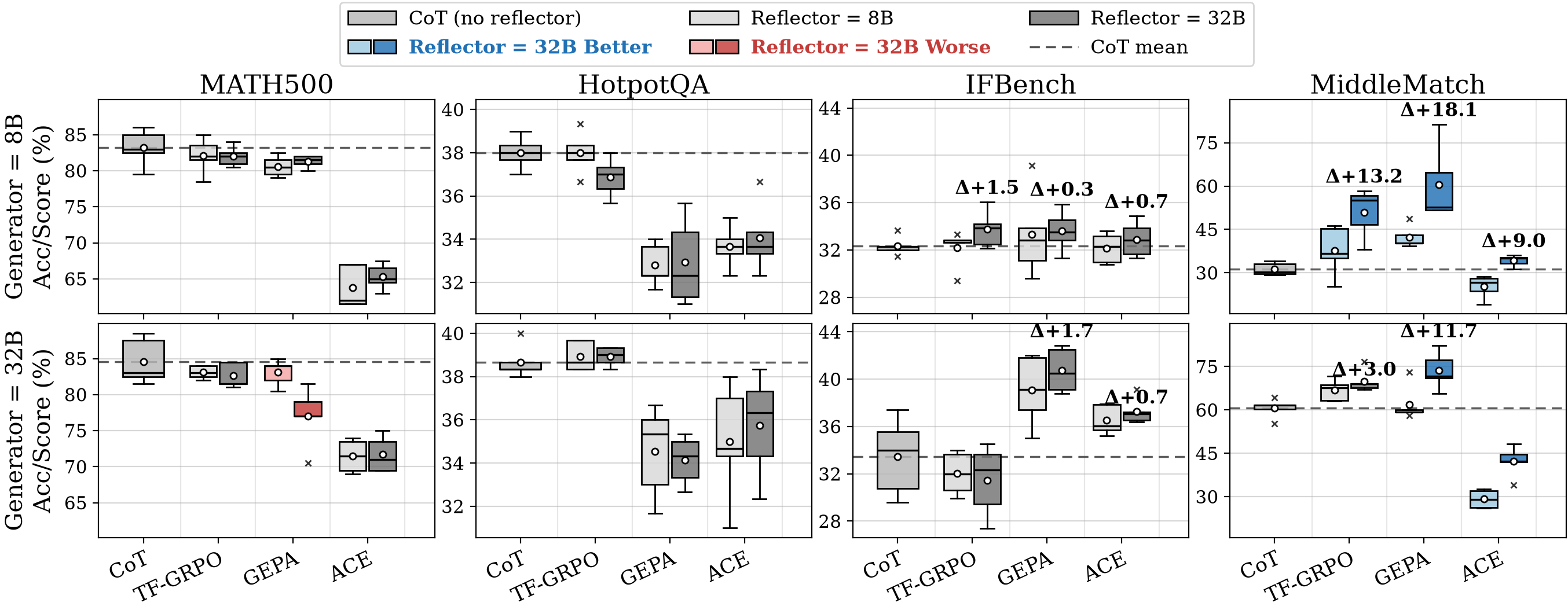}
    \vspace{-0.2in}
    \caption{Ablation on the scale of the Reflector and Generator across four datasets. Scaling the reflector cannot rescue failures but further enhances existing improvements.
    }
    % \junyuan{only use color if there is a significant difference.}\jr{done}}
    \label{fig:boxplot_results}
    \vspace{-0.15in}
\end{figure*}
After testing LCE methods with 8B-scale models, we further investigate whether scaling up models can improve the effectiveness of LCEs.
\\
\emph{(1) Bigger Reflector Can Cause More Gain but Also More Loss.}
In~\cref{fig:boxplot_results}, we investigate whether larger LLMs can engineer better context for smaller models or vice versa.
We define the {Reflector} as the model that generates reflections to engineer the context, and the {Generator} as the model that executes tasks using the optimized context.
Then, we conduct experiments on four datasets using GEPA, TF-GRPO, and ACE, varying the sizes of the Reflector and Generator between 8B and 32B. 
% On MATH500 and HotpotQA, performance gains from LCEs are marginal or negative across all combinations. 
We observe that although larger LLMs engineer more effective context in most settings on IFBench and MiddleMatch, they can even perform worse than smaller models on MATH500 and HotpotQA. 
% More observations are provided in Appendix~\ref{appendix:scaling_law}.
\\
\emph{(2) Bigger/Newer Models Do Not Necessarily Enable LCE Better Than Smaller/Older Ones.}
% \jr{}
% In addition to scaling the reflector and generator separately, we jointly scale the self-reflecting model from Qwen3-0.6B to Qwen3-235B-A22B on four datasets.
% \Cref{fig:gepa_scaling} shows that model scale primarily strengthens the underlying task solver, while the benefit of LCEs is inconsistent.
% On MiddleMatch, both GEPA and TF-GRPO improve over CoT across model sizes, with especially large gains at 32B.
% On IFBench, both methods match CoT at 235B size but cannot engineer useful context with fewer parameters.
% In contrast, neither method yields a reliable gain on MATH500 or HotpotQA: improvements at some smaller scales disappear or reverse as the model grows.
% Thus, more parameters are insufficient to promote LCEs to achieve better performance.
In addition to scaling the reflector and generator separately, we jointly scale the self-reflecting model from Qwen3-0.6B to Qwen3-235B-A22B across four datasets~\citep{Qwen3}.
\cref{fig:gepa_scaling} shows that model scaling primarily strengthens the underlying task solver, while the effectiveness of LCEs remain inconsistent.
On MiddleMatch, both GEPA and TF-GRPO outperform CoT across model sizes, with particularly large gains at 235B.
On IFBench, both methods match CoT at the 235B scale but fail to engineer useful context at smaller scales.
In contrast, neither method yields reliable gains on MATH500 or HotpotQA: improvements at some smaller scales disappear or even reverse as the model grows.
Thus, more parameters are insufficient to make LCEs consistently effective.

We further investigate whether newer LLM generations lead to better LCE behavior.
Specifically, we evaluate GEPA and TF-GRPO across five successive Qwen generations of comparable size, from Qwen1.5-7B to Qwen3.5-9B~\citep{qwen_series}, on four datasets.
As shown in \cref{fig:gepa_scaling}, although baseline performance generally improves across generations, the gains from LCEs still exhibit no consistent trend.
For example, GEPA shows smaller gains on HotpotQA as the LLM generation advances, while it continues to show substantial gains on MiddleMatch.
Overall, these results provide no clear evidence that newer LLM generations systematically lead to better LCE behavior.

% \begin{figure*}[h]
%     \centering
%     \vspace{-0.2in}
%     \includegraphics[width=\textwidth]{figs/scaling/gen_twocolors.png}
%     \vspace{-0.1in}
%     \caption{Generations scaling of self-reflection in LCEs from Qwen1.5-7B to Qwen3.5-9B across different datasets. CoT and GEPA initial are the initials for TF-GRPO and GEPA, respectively.}
%     \vspace{-0.2in}
%     \label{fig:generations_scaling}
% \end{figure*}

% \begin{takeawaybox}[colback=gray!5!white, colframe=gray!75!black,
% title=\ Key Takeaways]
% \ Scaling helps, but it depends on tasks.
% \ \begin{itemize}[leftmargin=*,nosep]
%   \item On tasks where rollouts lack useful knowledge (e.g., MATH500), scaling up the reflector fails to inject guidance and may even hurt performance.
%   \item On tasks where a small reflector already works well (e.g., MiddleMatch), a larger reflector produces better experiences and drives consistent gains for the same generator.
%   \item When a larger model has untapped potential (e.g., the 32B's high-variance CoT on IFBench), even a small reflector can effectively improve performance.
% \end{itemize}
% \end{takeawaybox}

\section{When \& How Can LLMs Engineer Effective Context?}

In addition to the aggregated \cref{tab:benchmark_all_models}, we will conduct qualitative and quantitative analysis to uncover the failure/success mechanisms of the LCEs.

% \junyuan{OUTLINE: The case study shows some potential success patterns, but it remains mysterious what is the general inherent cause of the success/failure. We notice the instruction adherence is generally connected with the same simple patterns (TODO: For example...).}
\par\noindent
\textbf{\underline{When Do LCEs Work?}} 
In \cref{tab:benchmark_all_models}, which method is effective is not simple.
We notice the largest gain (best LCE versus best baseline) in MiddleMatch, and the smallest gain in MATH500. In Ministral, LCEs are more useful than in Qwen3. Thus, we suppose both the data and the model play roles here. The intuition is that the LCEs learn from the model's failed rollouts and therefore the rollouts hinge not just on data but also on models. Specifically, we hypothesize that \emph{LCEs are more effective when rollouts exhibit homogeneous failure modes}.
To examine this hypothesis, we analyze the correlation between failure-mode diversity in the initial rollouts and performance gains \emph{across all experimental settings} (model, dataset, methods).
We characterize failure-mode diversity through the following steps:
\\
\noindent\emph{(1) Embedding Rollout Failure-modes.}
For each failed reasoning rollout, we diagnose its cause by using an LLM.
We instruct an LLM to summarize the cause mode, grounded in specific evidence and expressed in generalizable terms without instance-specific details (see exact prompt at~\cref{fig:failure_mode_prompt}).
% By default, we use the reasoning model itself to label the failure mode.
% Given a failed trajectory, we prompt an LLM to diagnose the trajectory and generate a failure mode $t_i$, a brief summary of failure reasoning.
% We ask the LLM to generate the mode, grounded in specific evidence and expressed in generalizable terms without instance-specific details (see exact prompt at~\cref{fig:failure_mode_prompt}).
We then encode each failure mode as a vector $\vx$ using the embedding model \texttt{all-mpnet-base-v2}~\citep{song2020mpnet}. The embedding model is widely used for short text; for example, ACE uses it to detect duplicate items in the playbook.
% where $d=768$ is the default output dimension of
% Here, each failure mode describes a categorized root cause of failure, grounded in specific evidence and expressed in generalizable terms without instance-specific details.
% We then encode each failure mode as a semantic embedding $\vx$
% where $d=768$ is the default output dimension of the embedding model \texttt{all-mpnet-base-v2}~\citep{song2020mpnet} which is widely used for embedding short text, for example, ACE uses it to detect duplicate items in the playbook.
\\
\noindent\emph{(2) Measuring Diversity.}
% We stack the embeddings $\vx_i$ as columns 
% \junyuan{rows instead of columns? Are you sure? Double check.} \junyuan{then $v\in \RR^{N\times 1}$ It does not make sense!} \jr{done}
% of a matrix $\mathbf{X}\in\mathbb{R}^{d\times N}$ and perform eigendecomposition of its Gram matrix 
% We use the Vendi Score~\citep{vendiscore}, which lies in $[1,N]$, to measure the failure-mode diversity. This metric can be interpreted as the effective number of distinct failure modes among the $N$ samples, accounting for their semantic similarities and relative frequencies, in which samller Vendi Score represents the more homogoues diversity.
We randomly subsample $N$ extracted failure-mode text samples and use the Vendi Score~\citep{vendiscore} to quantify failure-mode diversity. 
This score has been used to quantify pattern diversity, such as the diversity of generated images~\citep{vendi_example}.
The score ranges from $1$ to $N$ and can be interpreted as the effective number of distinct samples. 
The lower scores indicate more homogeneous failure modes.
Detailed experimental setup is in~\cref{appendix:failure_mode}.

In \cref{fig:model_38}, we present the performance gains versus Vendi scores and the mean/standard deviations of them on the axes.
Performance gain is defined as the performance score difference between after and before LCE.
We can see that different datasets present different diversity and the variance comes from models and LCE initial prompts.
MiddleMatch is the least diverse task, and AppWorld is the most diverse given that its task is to handle interactions with an app.

\begin{wrapfigure}{R}{0.62\linewidth}
    \centering
    \includegraphics[height=0.49\linewidth]{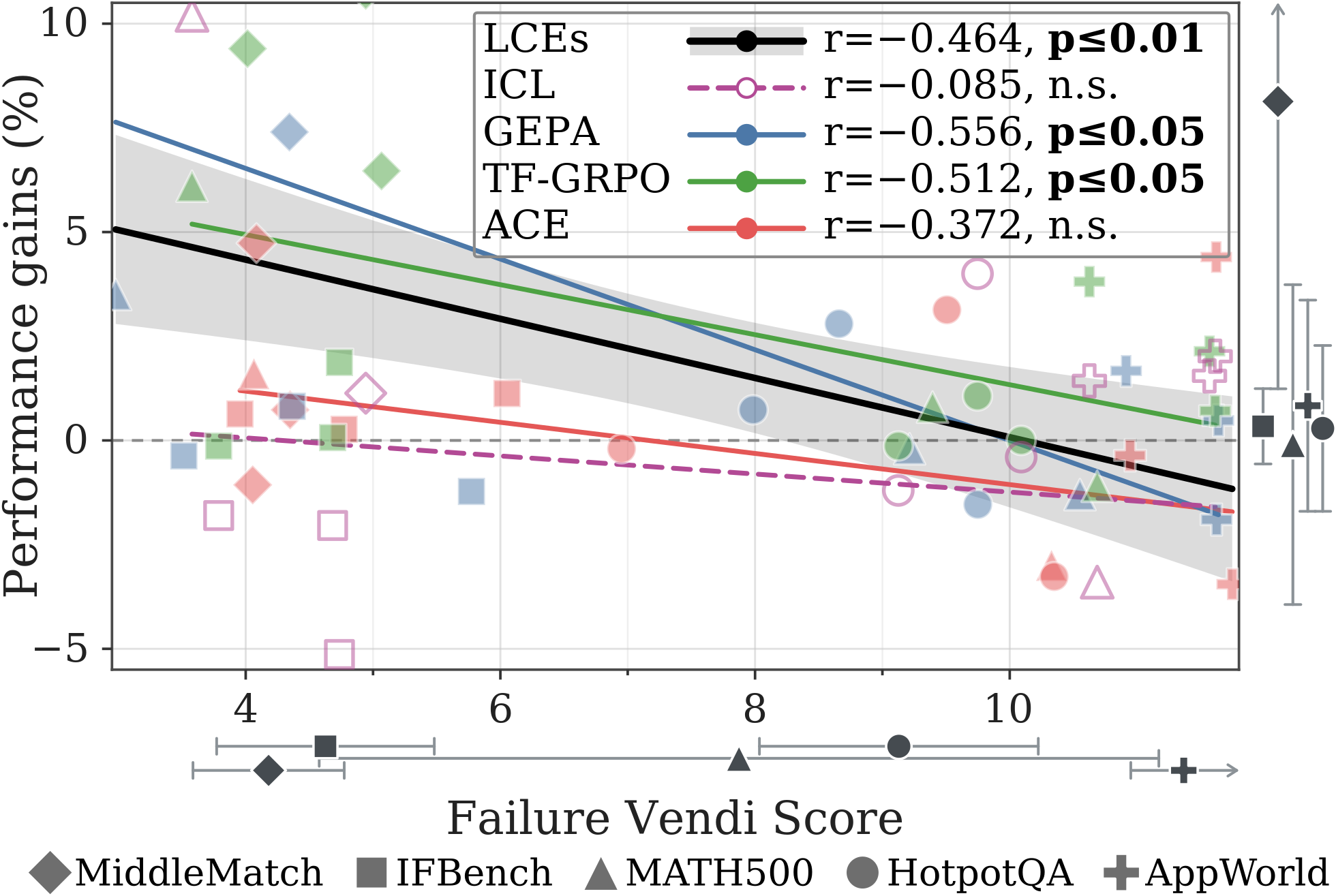}\hfill
    \includegraphics[height=0.49\linewidth]{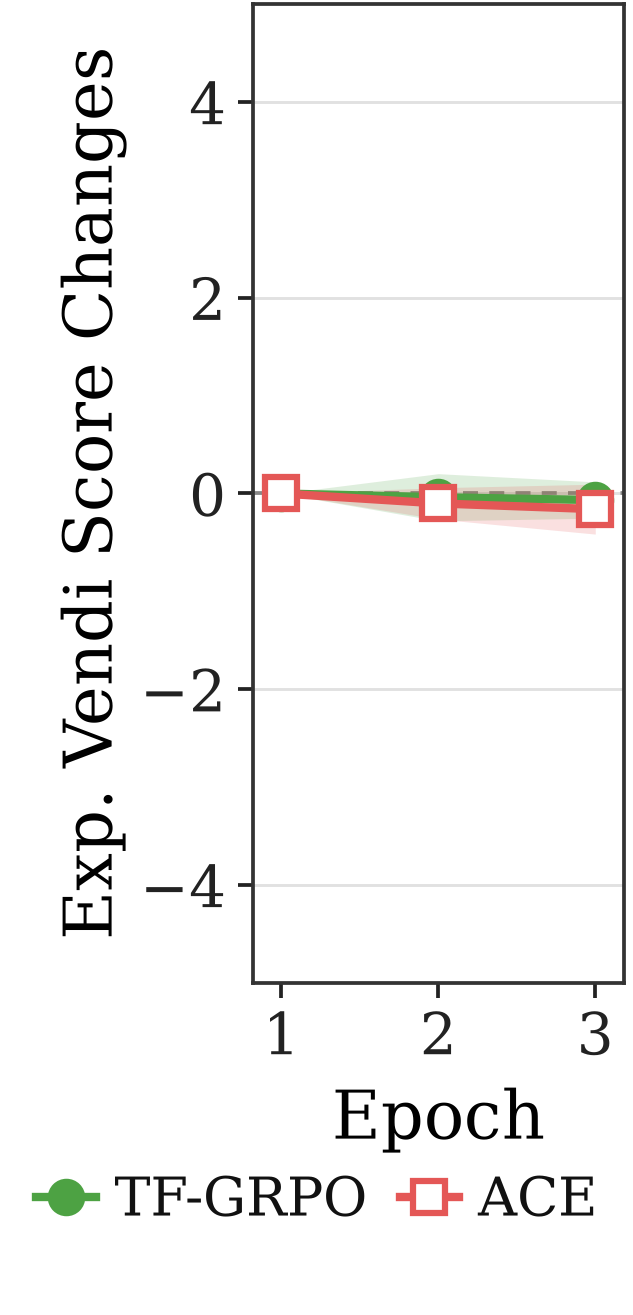}
    \caption{
    \textbf{Left:} Correlation between LCE performance gains
    and the \emph{Vendi Score of failure modes}.
    Markers and error bars show means and standard deviations on both axes for LCEs.
    The significant negative correlation indicates that
    LCEs tend to achieve larger performance gains when
    failure modes are less diverse. \textbf{Right:} Changes in the \emph{Vendi Score of experiences} across epochs. The small changes imply that LCEs struggle to distill experiences into shared principles.
    }
    \label{fig:model_38}
\end{wrapfigure}

Our major findings are:
\\
\emph{(1) Less Diverse Failure Modes, Better The LCEs Work.} 
The left figure clearly shows the negative correlation between diversity and performance gains across all LCE methods.
GEPA and TF-GRPO present the statistical significance, but ACE does not.
We conjecture that differences in context updates may explain this pattern. GEPA rewrites prompts, while TF-GRPO modifies or merges selected experiences. These updates may lose useful knowledge when failure modes are diverse. ACE’s incremental accumulation may better preserve such knowledge, making its performance gains less dependent on failure-mode diversity. 
\rev{To further control for the potential influence of other factors, we conduct a mixed-effects analysis in~\cref{app:mixed_effects}. The negative association between failure-mode diversity and performance gains remains significant, which is consistent with the above observations.}
\\
\emph{(2)} In contrast, the ICL gains turn out to be irrelevant to the diversity. That means the relation hinges on the LLM operations, for example, reflection on failures, which are missing from ICL.
\\
\emph{(3)} In \cref{fig:model_38} (Right), we show how the Vendi Score of experiences (e.g., distilled insights or a bulleted playbook) extracted by ACE and TF-GRPO from training rollouts changes over epochs across different models and datasets. 
The Vendi Score changes very mildly across iterations. That means the LLMs are not able to reduce the diversity of experiences by iteratively optimizing context, which explains why LCEs do not perform well on highly diverse tasks.
\\
\rev{
\textit{(4)} Beyond the above findings, we also conduct further analyses of potential factors influencing LCE effectiveness in~\cref{app:nonsig}. Factors, like optimization budgets across LCEs and differences in baseline task accuracy, do not show significant correlations with the performance gains.}

\begin{wrapfigure}{r}{0.35\linewidth}
    \centering
    \vspace{-0.1in}
    \includegraphics[width=\linewidth]
        {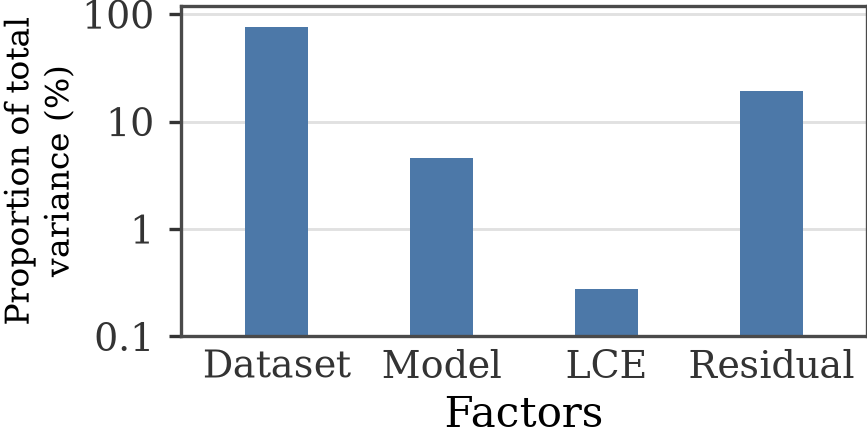}
    \vspace{-0.25in}
    \caption{Ratio of total variance in Vendi Score of failure modes.  
    % \junyuan{Change legned: Unified labeling, self-labeling. Reduce the width of the figure as small as possible.}
    % The dataset main effect accounts for the largest proportion of total variation, indicating that failure-mode diversity mainly comes from the data.
    % \junyuan{only use unified labeling. And no need to specify how you label here.}
    } 
    \vspace{-0.15in}
    \label{fig:effect_factors}
\end{wrapfigure}

% \textbf{LCEs cannot Reduce Experience Diversity by Iterations.} 
% \textbf{LCEs Cannot Reduce Experience Diversity by Iterations.} 
% The vendi score almost does not change by iterations.
% That means the LLMs are not able reduce the diversity of experiences by iteratively optimizing context.
% This explains why LCEs does not well on high-diverse tasks,

% \begin{wrapfigure}{r}{0.48\linewidth}
%     \centering
%     \vspace{-0.15in}
%     \includegraphics[width=\linewidth]
%         {figs/analysis/tfgrpo_experience_entropy_vs_epoch.png}
%     \caption{
%     Correlation between experience entropy and training epochs. Small $|r|$ and non-significant results imply that LCEs struggle to aggregate experiences into shared principles. 
%     \junyuan{Reduce width to 0.12 of linewidth, and put into \cref{fig:model_same_p_a} in same row. Share the caption. No need to show $|r|$. }
%     \junyuan{Change to connected lines. And compute the entropy w.r.t. the first epoch. We only care about changes.}
%     }
%     \vspace{-0.2in}
%     \label{fig:exps-p}
% \end{wrapfigure}

\par\noindent
\textbf{Failure-mode Diversity Mainly Comes from Data.}
To identify the main sources of variation in failure-mode diversity across settings, we decompose the variance of Vendi Scores into dataset, model, and LCE main effects, together with residual variance. 
In~\cref{fig:effect_factors}, we show the variance decomposition proportions. A larger variance proportion for a factor indicates greater differences in failure-mode diversity across its categories. 
The calculation step of the variance decomposition is provided in Appendix~\ref{appendix:calculation}. 
In the figure, the small variance proportion attributed to the model suggests that failure-mode diversity varies relatively little across models, whereas the large proportion attributed to the dataset indicates substantial variation across datasets. This suggests that failure-mode diversity is driven primarily by the dataset, with some datasets naturally eliciting a broader range of failure modes.
For example, the high similarity in MiddleMatch queries contributes to its low failure diversity: In MiddleMatch, questions are generated using a fixed template: \emph{``Here's a list of names: \texttt{\textbackslash n\textbackslash n}\{sequence of names\}\texttt{\textbackslash n\textbackslash n} What is the single name that appears right between \texttt{<name\_1>} and \texttt{<name\_2>}?''}
Such homogeneous data patterns make the reflection complexity lower and therefore lower failure mode diversity.

\par\noindent
\textbf{\underline{How Do LCEs Work?}}
We have seen that LCEs work well on low-diversity datasets but it also happens, though not frequently, that LCEs can sometimes improve performance on high-diversity tasks. For example, in MATH500 with Gemma4 and HotpotQA with Min3, LCEs exhibit high variance in both performance gains and failure-mode Vendi Scores (~\cref{fig:model_38}). Here, we conduct case studies to understand how LCEs conquer these difficult tasks.

\begin{figure*}[ht]
    \centering
    % \vspace{-0.1in}
    \begin{minipage}[t]{0.49\textwidth}

        \centering
        \includegraphics[width=\linewidth]
            {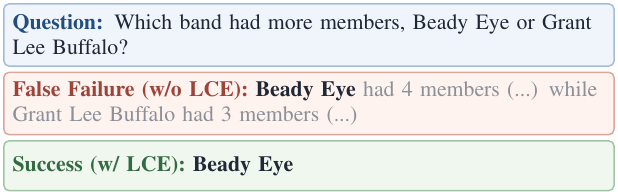}
    
        \includegraphics[width=\linewidth]
            {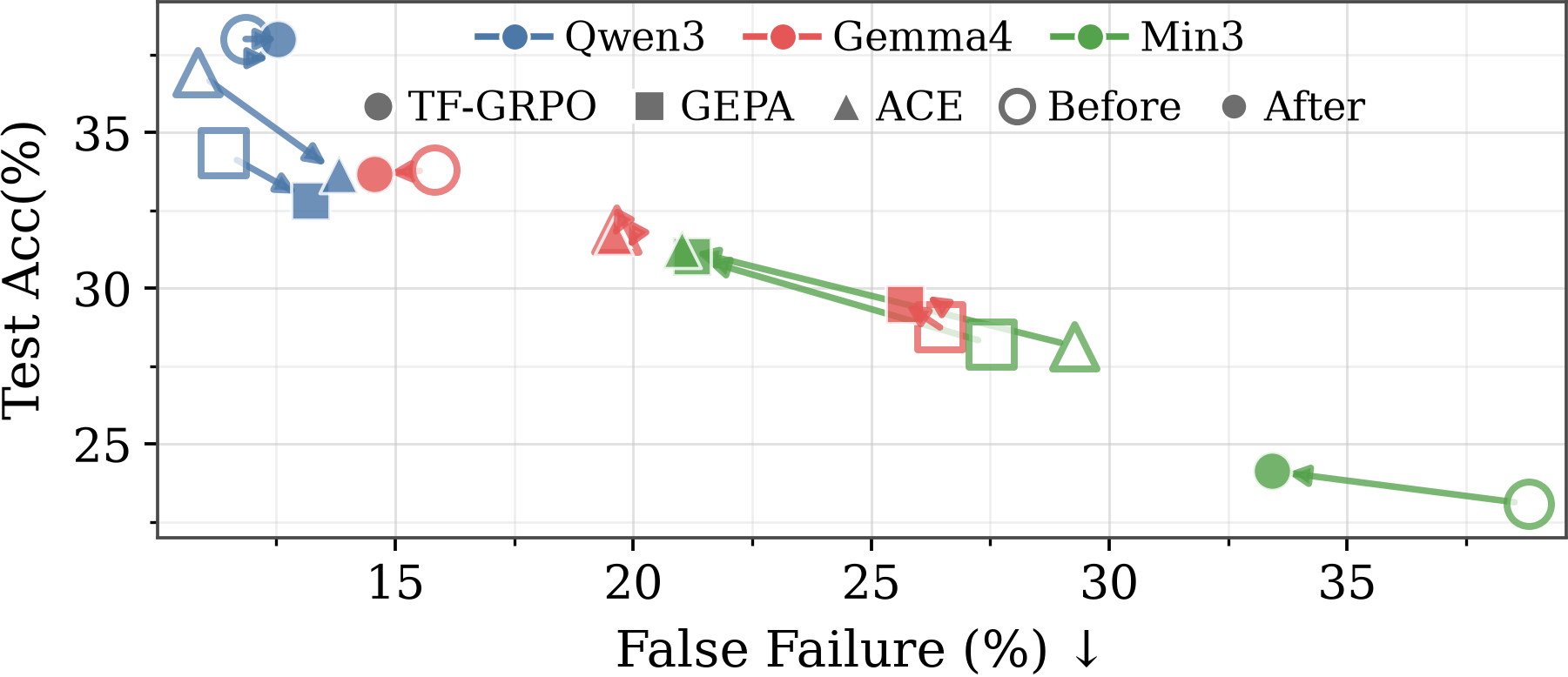}
        \vspace{-0.2in}
        \captionof{figure}{LCE behavior on HotpotQA. \textbf{Top:} Example of a verbose prediction from Min3 leading to failure. \textbf{Bottom:} LCEs reduce the false failure rate in Ministral alongside performance gains.}
        \vspace{-0.1in}
        \label{fig:lce_hotpotqa}
    \end{minipage}
    \hfill
    % 右图：MATH500
    \begin{minipage}[t]{0.49\textwidth}
        \centering
        \includegraphics[width=\linewidth]
            {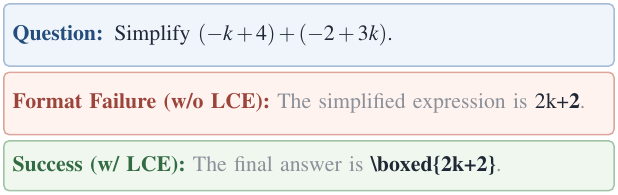}

        \includegraphics[width=\linewidth]
            {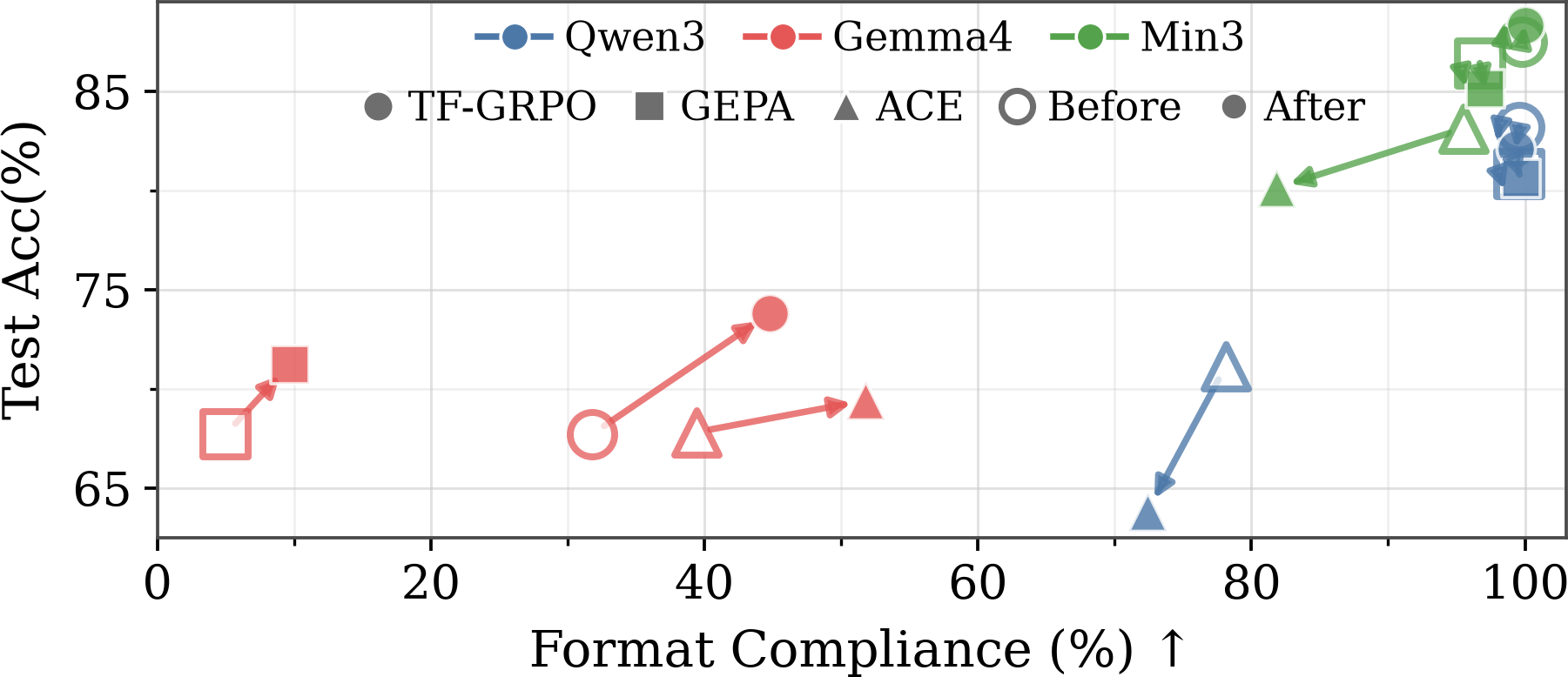}
        \vspace{-0.2in}
        \captionof{figure}{LCE behavior on MATH500. \textbf{Top:} Example of a format failure from Gemma4. \textbf{Bottom:} LCEs improve compliance with the required format in Gemma4 while improving performance.}
        \vspace{-0.1in}
        \label{fig:lce_math500}
    \end{minipage}
    % \vspace{-0.1in}
\end{figure*}

\begin{table*}[t]
    % \vspace{-0.05in}
    \caption{Performance under different experience utilization and retrieval
    settings. \textbf{R}, \textbf{E}, and $\boldsymbol{\phi}$ denote training
    rollouts, knowledge extraction, and experience consolidation, respectively.
    $k$ denotes the number of examples or experiences used, with \textcolor{gray}{var.} indicating that the number varies depending on the optimized experiences. \textbf{Bold} and \underline{underlined} values indicate the best and second-best results.
    Q, G, and M denote Qwen3, Gemma4, and Min3, respectively. 
    }
    \label{tab:grpo-ablation}
    \vspace{-0.05in}
    \footnotesize
    \centering
    \setlength{\tabcolsep}{3pt}
    \renewcommand{\arraystretch}{1.1}
    \begin{tabular}{@{}lcccc|cccc@{}}
    \toprule

    \multirow{2}{*}{\textbf{Method}}
    & \multirow{2}{*}{\textbf{R}}
    & \multirow{2}{*}{\textbf{E}}
    & \multirow{2}{*}{$\boldsymbol{\phi}$}
    & \multirow{2}{*}{$k$}
    & \textbf{MiddleMatch}$^{\mathrm{Q}}$
    & \textbf{MATH500}$^{\mathrm{G}}$
    & \textbf{MATH500}$^{\mathrm{M}}$
    & \textbf{IFBench}$^{\mathrm{G}}$ \\

    \cmidrule(lr){6-7}
    \cmidrule(lr){8-9}

    & & & &
    & \multicolumn{2}{c}{$\mathrm{ICL}>\mathrm{LCE}>\mathrm{CoT}$}
    & \multicolumn{2}{c}{$\mathrm{LCE}\geq\mathrm{CoT}>\mathrm{ICL}$} \\

    \midrule

    CoT
    & & & & 0
    & $31.27{\scriptscriptstyle\pm2.10}$
    & $67.70{\scriptscriptstyle\pm1.04}$
    & $\underline{87.50}{\scriptscriptstyle\pm1.06}$
    & $39.76{\scriptscriptstyle\pm2.22}$ \\

    ICL
    & \checkmark & & & 3
    & $\mathbf{46.07}{\scriptscriptstyle\pm4.80}$
    & $\mathbf{77.90}{\scriptscriptstyle\pm3.54}$
    & $68.10{\scriptscriptstyle\pm2.41}$
    & $34.63{\scriptscriptstyle\pm2.17}$ \\

    \midrule

    GRPO-ICL RAG
    & \checkmark & & & 3
    & $\underline{44.00}{\scriptscriptstyle\pm3.46}$
    & $\underline{74.10}{\scriptscriptstyle\pm2.25}$
    & $26.50{\scriptscriptstyle\pm4.23}$
    & $37.41{\scriptscriptstyle\pm1.44}$ \\

    All-Exps Random
    & \checkmark & \checkmark & & 3
    & $32.00{\scriptscriptstyle\pm6.00}$
    & $69.70{\scriptscriptstyle\pm0.91}$
    & $87.40{\scriptscriptstyle\pm0.55}$
    & $\underline{40.92}{\scriptscriptstyle\pm2.07}$ \\

    % All-Exps RAG
    % & \checkmark & \checkmark & & 3
    % & $39.07{\scriptscriptstyle\pm11.87}$
    % & $68.30{\scriptscriptstyle\pm2.89}$
    % & $\mathbf{88.50}{\scriptscriptstyle\pm1.27}$
    % & $38.78{\scriptscriptstyle\pm1.67}$ \\

    % \midrule

    Final-Exps
    & \checkmark & \checkmark & \checkmark & \textcolor{gray}{var.}
    & $37.73{\scriptscriptstyle\pm8.57}$
    & $73.80{\scriptscriptstyle\pm4.12}$
    & $\textbf{88.30}{\scriptscriptstyle\pm1.15}$
    & $\mathbf{41.63}{\scriptscriptstyle\pm1.21}$ \\

    % Final-Exps RAG
    % & \checkmark & \checkmark & \checkmark & 3
    % & $38.53{\scriptscriptstyle\pm5.62}$
    % & $68.80{\scriptscriptstyle\pm1.04}$
    % & $87.90{\scriptscriptstyle\pm1.88}$
    % & $40.00{\scriptscriptstyle\pm2.11}$ \\

    \bottomrule
    \end{tabular}
    \vspace{-0.15in}
\end{table*}

\par\noindent
\textbf{Case Study: LCEs improve adherence to the task instructions.}
In multiple samples, we notice that the experiences extracted in LCEs are related to the output format or instruction details, less relevant to problem solution ideas. 
\textbf{(1)} In \emph{HotpotQA}, by investigating the generated samples, we find that Min3 tends to generate verbose answers for problems, which sometimes cannot be identified as an answer by the evaluation. One example is given in \cref{fig:lce_hotpotqa} (Top), where we show that both answers contained the correct answer (bolded) but LCE makes the model produce the recognizable answer.
For HotpotQA, the evaluation relies on exact matching of answers, and thus only the LCE answer is considered correct.
Beyond the example, we found that the improvement of LCEs on HotpotQA is related to the reduction of such ``False Failure''.
During optimization, LCEs repeatedly identify this verbosity failure and reinforce conciseness through instructions such as \emph{``Answer only what the question asks. Avoid extraneous details.''}
Concretely, in \cref{fig:lce_hotpotqa} (Bottom), we show that the accuracy increase is related to the reduced False Failure rate, i.e., the proportion of predictions marked as incorrect by Exact Match that, however, actually contain the ground-truth answer. 
\textbf{(2)} For \emph{MATH500}, we observe that Gemma4 frequently fails to follow the format requirement, such as 
\texttt{``Put the final answer in \textbackslash boxed\{...\}''}, as illustrated in \cref{fig:lce_math500} (Top). Thus, in \cref{fig:lce_math500} (Bottom), we report format compliance before and after optimization across models. We found that LCEs reinforce this instruction in the optimized context, improving format adherence and contributing to Gemma4's performance gains greatly.
Details on calculating the false failure rate and format compliance are in~\cref{appendix:case_study}. \rev{Moreover, we apply an LLM-based answer extractor to standardize final-answer extraction in~\cref{app:reeval} and find that the performance gains of LCEs decrease substantially after extraction, further suggesting that the improvements of LCEs come from the better instruction adherence.} 
% \junyuan{@JZ, please make sure format compliance implementation details are provided.} \jr{done}

\par\noindent
\textbf{\underline{Case Study: TF-GRPO.}} To further investigate method-level failures, we conduct an additional ablation study with TF-GRPO, whose framework follows the general principles in~\cref{eq:po} with clearly separated workflow stages and achieves the best performance in more settings in~\cref{tab:benchmark_all_models} across LCEs.
In \cref{tab:grpo-ablation}, we compare the ablations under two representative performance regimes based on whether LCE beats baselines: ICL $>$ LCE $>$ CoT and LCE $\geq$ CoT $>$ ICL.
% (1) Random-ICL $>$ LCE $>$ CoT: In-context demonstrations provide useful information that improves performance. Although LCEs also extract useful information, they underperform Random-ICL. 
% % We examine MiddleMatch with Qwen3 and MATH500 with Gemma4 under this regime.
% (2) LCE $\geq$ CoT $>$ Random-ICL: LCEs maintain or improve performance relative to CoT, whereas Random-ICL degrades it. 
% % We analyze IFBench with Gemma4 and MATH500 with Min3 to investigate this contrast.
Here, we add two inference variants that use a fixed number of experiences or in-context examples:
(1) GRPO-ICL RAG (retrieval-augmented generation) retrieves in-context demonstrations used during TF-GRPO training, excluding query samples that can be solved in all rollouts.
% excluding those from problems solved correctly in all attempts.
% (2) All-Exps RAG retrieves experiences from the full set of extractor-proposed experiences accumulated during TF-GRPO training.
(2) All-Exps Random randomly samples experiences from all experiences produced by TF-GRPO (Final-Exps).
% same pool used by All-Exps RAG.
% (4) Final-Exps RAG retrieves experiences from the final experience pool produced by TF-GRPO (Final-Exps).
The RAG variants retrieve items based on semantic similarity to the retrieval query using \texttt{Qwen3-Embedding-0.6B}~\citep{qwen3_embedding}, which supports encoding long in-context demonstrations.
Beyond the inference variants considered above, we also evaluate several additional variants. As these experiments offer limited new insights, we report the full results in~\cref{appendix:tfgrpo_case_study}.

\par\noindent
\textbf{Experience extraction (\textbf{E}) plays the most critical role.}
Across all four cases, either catastrophic performance loss (left two columns) or gain (right two columns) starts from the All-Exps Random, where Experience Extraction was first introduced to the ablation.
That means when LLMs did not extract useful experiences from rollouts, the performance will not be improved over simply ICL of rollouts.
Instead, when useful experiences are extracted (i.e., gains via \textbf{E}), the shorter experiences improve performance. We conjecture it is due to the smaller context length.
The conjecture can be supported by our observation in \cref{appendix:length}, that lengthy contexts can diminish the benefits of useful knowledge in ICL.

\section{Conclusion}
\label{sec:conclusion}
\vspace{-0.1in}

In this paper, we study LCEs across diverse tasks and LLM families. No method consistently outperforms even a simple CoT or ICL baseline, and larger or newer models do not reliably improve LCE effectiveness. We therefore conduct further experiments to investigate when and how LCEs are effective. We find that LCEs work better when facing less diverse failure modes, while such diversity varies substantially across datasets. Extracted experiences show no consistent reduction in diversity over training iterations, suggesting that current LCEs struggle to distill shared principles. Our ablations also show that converting training rollouts into experiences can discard useful information. Together, these findings identify limited extraction capability as a key bottleneck in turning diverse failures into broadly useful context.
\par\noindent
\textbf{Guidelines for future studies.}
Our work provides insights into the effectiveness and working mechanisms of LCEs. Yet, it is still open how to improve the LCEs.
Our insights suggest that it is important to test methods under diverse models and datasets to fully demonstrate their generalization.
LCE method development should pay more attention to overcoming the bottleneck in extracting useful experiences from diverse failure modes of reasoning.
Moreover, future work may also explore incorporating knowledge on demand to reduce irrelevant content and control context length as the training data grows.

\section{Acknowledgements}
This research/project is partially supported by the National University of Singapore under the NUS Start-Up Grants (FY2026).

\bibliographystyle{costa_preprint}
\bibliography{reference}

\clearpage
\appendix
% \jr{I have not revised the appendix currently}

\section{Evaluation Protocols in Representative Methods}

\begin{table}[h]
  \centering
  \caption{{Reported benchmarks of evaluated context engineering methods.}}
  \vspace{-0.1in}
  \label{tab:dataset_coverage}

  \small
  \setlength{\tabcolsep}{4pt}
  \renewcommand{\arraystretch}{1.02}
  \setlength{\extrarowheight}{1pt}
  \setlength{\aboverulesep}{0pt}
  \setlength{\belowrulesep}{0pt}
  \renewcommand{\tabularxcolumn}[1]{m{#1}}

  \begin{tabularx}{\linewidth}{
    >{\raggedright\arraybackslash}m{0.14\linewidth}
    >{\centering\arraybackslash}X
  }
    \toprule
    \textbf{Method}
    & \textbf{Reported Benchmarks} \\[2pt]
    \midrule

    % MIPROv2
    % & HotpotQA, HoVer, Iris, ScoNe, Heart~Disease
    %   ~\citep{HotpotQA,HoVer,Iris,scone,heart_disease}
    %   \\[2pt]

    TextGrad
    & BBH, GSM8K~\citep{BBH,gsm8k}
      \\[2pt]

    GEPA
    & HotpotQA, IFBench, HoVer, PUPA, AIME25, LiveBench-Math
      ~\citep{HotpotQA,HoVer,IFbench,PUPA,livebench}
      \\[2pt]

    TF-GRPO
    & AIME24, AIME25, WebWalkerQA
      ~\citep{AIME,webwalker}
      \\[2pt]

    ACE
    & AppWorld, FiNER, Formula~\citep{AppWorld,FiNER,Formula}
      \\
    \bottomrule
  \end{tabularx}
  \vspace{-0.15in}
\end{table}

\section{Experimental Setup} 

\subsection{Experimental details for Benchmark and Scaling Experiments}
\label{appendix:setup}
We use Qwen3-8B~\citep{Qwen3}, Gemma-4-E4B-it~\citep{team2026gemma}, and Ministral-3-8B-Instruct-2512-BF16~\citep{liu2026ministral} for the main benchmark. Our scaling experiments examine both model size, from Qwen3-0.6B to Qwen3-235B-A22B, and model generation, from Qwen1.5-7B to Qwen3.5-9B.
For each task, we use the same configuration across all models and LCEs in both the main benchmark and the scaling experiments. For single-step experiments, we set the maximum output length to $16{,}384$ tokens and the sampling temperature to $T = 0.7$. For AppWorld, we use the same sampling temperature and a maximum output length of $4{,}096$ tokens per step.

% \junyuan{What is the model? Qwen3 8B by default?}
\par\noindent
\textbf{\underline{Methods.}} We adopt the official codebase of each method in our single-step experiment.
% MIPROv2~\citep{MIPROv2} and 
GEPA~\citep{GEPA} is officially supported in the DSPy package~\citep{DSPy}, and we set \texttt{auto="heavy"} to maximize performance.
For TextGrad~\citep{TextGrad}, we use the official package with 3 epochs and a batch size of 5.
For TF-GRPO~\citep{TFGRPO}, we follow the official setting with 3 epochs and a group size of 5, using the original prompt templates for the critic and summarizer; to align with other methods and maximize performance, we run validation after each epoch and select the best experience pool.
For ACE~\citep{ACE}, we use the official implementation with a curator frequency of 5 and 3 epochs with bullet de-duplication.
For AppWorld, we use the ACE and GEPA implementations provided in the official ACE codebase and adapt TF-GRPO to the same framework with the original prompts. We set the length limits for per-step execution outputs and the total message history to $10{,}000$ and $200{,}000$ characters, respectively. 
For ACE on AppWorld, we use the default settings: a curator frequency of 1 and a maximum of 5 Reflector refinement rounds.
Each method is run with 5 random seeds ({\{0, 1, 2, 3, 4\}}). All models are served with vLLM on NVIDIA GPUs.

% Each optimization method is run 5 times with different seeds\junyuan{what are the seed values?}. All models are served with vLLM on NVIDIA GPUs.

% \junyuan{Please move all exp setting details from Sec 3 to here.}
% \junyuan{moved from TF-GRPO. Need to rewrite.} We follow the official setting with epoch set to 3 and group size set to 5, using the same prompt templates for critic and summarizer~\citep{TFGRPO}. Additionally, to align with other methods and maximize performance, we run validation after each epoch to select the best experience pool.
% \junyuan{moveed from ACE. Need to rewrite.}
% We use the official implementation~\citep{ACE}, and to avoid an overly long playbook, we set the batch size to 5 and the epoch number to 3.
% % \noindent\textbf{\underline{Configuration}}
% Each optimization method is run 5 times with different seeds\junyuan{what are the seed values?}. All models are served with vLLM on NVIDIA GPUs.
\par\noindent
\textbf{\underline{Benchmarks and Datasets.}}
In \cref{tab:dataset_coverage}, we show that existing methods are mostly evaluated on disjoint datasets.
% \junyuan{I added this paragraph to highlight the difference between tasks in the view of capabilities.}
To understand the landscape, we select a variety of datasets with different reasoning patterns: \emph{Reasoning} to find the solution (MATH500~\citep{MATH500}, HotpotQA~\citep{HotpotQA}); \emph{Retrieval} from long context (HotpotQA~\citep{HotpotQA}, MiddleMatch~\citep{MiddleMatch}); \emph{Following Instructions} with unseen constraints (IFBench~\citep{IFbench}) and \emph{Agentic Tasks} requiring multi-step planning and execution (AppWorld~\citep{AppWorld}).
In these tasks, training-based methods like RLHF/RLVF have been shown to significantly improve LLMs' capabilities \citep{RLHF,PPO,DPO}. But little work was done to examine the LCEs in these tasks in a unified setting.
% Math and HotpotQA are reasoning-intensive tasks. Math relies on the LLM's capability to reason and solve a hard problem, while HotpotQA requires reasoning with retrieval from context. IFBench tests if the LLM can precisely follow instructions with out-of-domain constraints. MiddleMatch is a retrieval-focused task to find value associated with the anchor name.
% \junyuan{In these tasks, do training methods significantly improve models?}
\\
\noindent\textbf{MATH500}~\citep{MATH500} is a 500-problem subset of competition-level mathematics drawn from the MATH benchmark~\citep{MATH}. Following prior settings, we use a 100/200/200 split for training/validation/test, and adopt Qwen-Verify to grade the results.
\\
\noindent\textbf{HotpotQA}~\citep{HotpotQA} is a multi-hop question-answering dataset in which the model must aggregate evidence across multiple documents. To unify the training framework across methods, we follow prior settings~\citep{TextGrad} and provide the gold and distractor passages directly in the context, thereby removing the retrieval stage. Following GEPA~\citep{GEPA}, we use a 150/300/300 split and adopt exact match as the evaluation metric.
\\
\noindent\textbf{IFBench}~\citep{IFbench} is an instruction-following dataset designed to evaluate whether LLMs can comply with user instructions rather than merely answer the underlying question. 
% Unlike other datasets, IFBench's test set mainly targets out-of-distribution generalization. 
Following GEPA~\citep{GEPA}, we use a 150/300/294 split and adopt the official IFBench~\citep{IFbench} metric.
\\
\noindent\textbf{MiddleMatch}~\citep{MiddleMatch} is a basic retrieval benchmark in which the model must identify the target name positioned between two given anchor names. We generate the data following the procedure of prior work, and use a 100/100/300 split with exact match as the evaluation metric.
\\
\noindent\textbf{AppWorld}~\citep{AppWorld} is a benchmark of stateful tasks spanning multiple applications, where agents interpret user requests and interact with applications through APIs.
We use the standard 90/57/168 training/development/test-normal split and adopt the official task goal completion as the evaluation metric.

% \subsection{Experimental Setup}

% \noindent\textbf{\underline{Prompt Templates}}
% We make minor adaptations to each method's prompt to follow its original design while keeping the core instructions consistent across methods. Since the resulting prompts differ across methods and are therefore not directly comparable, we report each template's pre-optimization performance separately. Full prompt templates are provided in Appendix~\ref{subsec:prompt-template}.
\par\noindent
\underline{\textbf{Prompt Templates}}
\label{subsec:prompt-template}
We make minor adaptations to each method's prompt to follow its original design while keeping the core instructions consistent across methods. Since the resulting prompts differ across methods and are therefore not directly comparable, we report each template's pre-optimization performance separately. 
Here we use MATH500 as a running example to introduce the fully initial prompt template of different methods.

\begingroup
\captionsetup[figure]{position=bottom, skip=4pt}
\begin{lstlisting}[style=promptstyle, belowskip=0pt]
Variables: Solve the following problem step by step. %%% will add specific prompt according to the datasets

Fixed: Put your final answer within \\boxed{{}}. %%% will be changed to other required format instructions according to the datasets
\end{lstlisting}
\captionof{figure}{Prompt Template of TextGrad}
% \label{fig:prompt_textgrad}
\endgroup

\begingroup
\captionsetup[figure]{position=bottom, skip=4pt}
\begin{lstlisting}[style=promptstyle, belowskip=0pt]
Solve the following problem step by step. Put your final answer within \\boxed{{}}. %%% will be changed to task prompt with format instructions
{problem}

When solving problems, you MUST first carefully read and understand the helpful instructions and experiences:
{experiences}
\end{lstlisting}
\captionof{figure}{Prompt Template of TF-GRPO.}
% \label{fig:prompt_textgrad}
\endgroup

\begingroup
\captionsetup[figure]{position=bottom, skip=4pt}
\begin{lstlisting}[style=promptstyle, belowskip=0pt]
Module: dspy.ChainOfThought(CoTSignature)

class CoTSignature(dspy.Signature):
    question = dspy.InputField()
    answer = dspy.OutputField(desc="Put your final answer within \\boxed{{}}.") %%% desc will be changed to format instructions

Variables: Given the fields `question`, produce the fields `answer`. %%% will be changed to task prompt
\end{lstlisting}
\captionof{figure}{Prompt Template of GEPA.}
% \label{fig:prompt_textgrad}
\endgroup

\begingroup
\captionsetup[figure]{position=bottom, skip=4pt}
\begin{lstlisting}[style=promptstyle, belowskip=0pt]
You are an analysis expert tasked with answering questions using your knowledge, a curated playbook of strategies and insights and a reflection that goes over the diagnosis of all previous mistakes made while answering the question.

**Instructions:**
- Read the playbook carefully and apply relevant strategies, formulas, and insights
- Pay attention to common mistakes listed in the playbook and avoid them
- Show your reasoning step-by-step
- Be concise but thorough in your analysis
- If the playbook contains relevant code snippets or formulas, use them appropriately
- Double-check your calculations and logic before providing the final answer

Your output should be a json object, which contains the following fields:
- reasoning: your chain of thought / reasoning / thinking process, detailed analysis and calculations
- bullet_ids: each line in the playbook has a bullet_id. all bulletpoints in the playbook that's relevant, helpful for you to answer this question, you should include their bullet_id in this list
- final_answer: put your final answer within \\boxed{{}}. %%% will be changed to task prompt with format instructions


**Playbook:**
{}

**Reflection:**
{}

**Question:**
{}

**Answer in this exact JSON format:**
{{
  "reasoning": "[Your chain of thought / reasoning / thinking process, detailed analysis and calculations]",  
  "bullet_ids": ["calc-00001", "fin-00002"],  
  "final_answer": "[Your concise final answer here]"
}}

---
\end{lstlisting}
\captionof{figure}{Prompt Template of ACE.}
% \label{fig:prompt_textgrad}
\endgroup

GEPA and TextGrad update the entire prompt by default, so we keep the format instructions fixed and optimize only the task prompt. 
For TextGrad and TF-GRPO, we append the task prompt and format requirements to the CoT prompt. 
Since we use the CoT module in GEPA, we replace the CoT.instructions in the DSPy module with the task prompt. ACE provides its own CoT template, so we only replace its final answer block with the task prompt and format requirements, as illustrated in our MATH500 example.
We list the task-specific prompts and format requirements for each dataset in~\cref{fig:task_prompt}. For HotpotQA, we use the default template and append additional format instructions following GEPA’s setup. For IFBench, we use the same instructions as GEPA. For MiddleMatch, we use the prompt from prior work~\citep{MiddleMatch} and add output-format requirements to extract the answer.

\begingroup
\captionsetup[figure]{position=bottom, skip=4pt}
\begin{lstlisting}[style=promptstyle, belowskip=0pt]
HotpotQA: 
Task prompt: [None]
Format instructions: Directly provide the answer inside <answer> and </answer>, without detailed illustrations, for example, <answer> Beijing </answer>.

MATH500: 
Task prompt: [None]
Format instructions: Put your final answer within \\boxed{{}}.

IFBench:
Task prompt: Respond to the query. Ensure the response is correct and adheres to the given constraints.
Format instructions: [None]

MiddleMatch:
Task prompt: [None]
Format instructions: Put the exact final name within \boxed{}, for example, \boxed{Michael Jackson}.
\end{lstlisting}
\captionof{figure}{Task prompt and format instructions for different tasks.}
\label{fig:task_prompt}
\endgroup

For AppWorld, we adopt the initial prompts for GEPA and ACE from the ACE implementation and use GEPA’s initial prompt for TF-GRPO.

% \section{Test Performance using Qwen3}
% \label{appendix:test_performance}
% \begin{figure*}[ht]
%     \centering
%     \setlength{\tabcolsep}{1pt}
%     \renewcommand{\arraystretch}{0}
%     \includegraphics[width=\textwidth]{figs/benchmark/boxplot_Q1.5.png}
%     \caption{Performance across LCEs on different benchmarks.}
%     \label{fig:pass_at_k}
% \end{figure*}

% For TextGrad, its poor performance in validation means it cannot engineer context in most settings; e.g., 5-seed context engineering on HotpotQA degrades to CoT prompting, which shows its limited capacity to fit on the training and validation sets. Following is an example of failure in context engineering on MATH500, the major reason is TextGrad overfits to the specific problem and hacks the LLM to force them only output on this selected problem rather than a real problem. 

\subsection{Experimental details for Failure-Mode Extraction}
% \section{Additional setup of Failure-Mode Extraction}
\label{appendix:failure_mode}
\textbf{\underline{Configurations}} We use Qwen3.8-27B~\citep{qwen38} to uniformly label the failure modes. 
For each experimental setting (model, dataset, method), we collect failed training trajectories generated using the initial context across five runs with different random seeds. We then combine these trajectories into a single set.
Since the Vendi Score depends on sample size, we randomly sample the same number of trajectories, $N$, for each experimental setting to calculate the score. We repeat this sampling and calculation 100 times for each setting and report the mean score.
\par\noindent
\textbf{\underline{Prompt Template for Failure Mode Extraction}} The prompt templates used to extract failure modes for different tasks are shown in~\cref{fig:failure_mode_prompt}.
% \label{appendix:failure_mode}
\begingroup
\captionsetup{type=figure, position=bottom, skip=4pt}
\captionsetup[subfigure]{position=bottom, skip=4pt}

\begingroup
\begin{lstlisting}[style=promptstyle, belowskip=0pt]
Diagnose the failure in this validated error trajectory.

Describe the smallest reusable failure mechanism that caused the error.

Requirements:
- Return exactly one short, specific sentence.
- Identify the failed operation, relation, constraint, evidence use or inference step.
- Use semantic roles instead of instance-specific names, dates, values, or positions.
- Do not add background, advice, headings, bullets, or generic evaluation.
- Do not state only the outcome, such as "incorrect answer" or "reasoning failure".
- Base the diagnosis only on the supplied trajectory and evaluator evidence.
- Failure may be caused by output that violates the task instructions, such as an unexpected format.

Question:
{question}

Task instruction used by the generator:
{task_instructions}

Input context: %%%Only provide for HotpotQA
{input_context}

Gold supporting-facts passages for only diagnosing evidence use: %%%Only provide for HotpotQA
{supporting_facts_text}

Reference answer: %%%Only provide for IFBench
{ground_truth}

Complete model reasoning or raw output:
{raw_output}

Extracted prediction following the task instructions:
{extracted_prediction}

Score:
{score}

Evaluator feedback: %%% Only provide for IFBench 
{feedback}

Return only JSON with exactly one field:
{"failure_mechanism": "One short sentence describing the reusable failure mechanism."}
\end{lstlisting}
\subcaption{Prompt for single-step experiments.}
\label{fig:prompt_a}
\endgroup

\begingroup
\begin{lstlisting}[
    style=promptstyle,
    aboveskip=6pt,
    belowskip=0pt
]
Diagnose the failure in this validated error trajectory.

Describe the smallest reusable failure mechanism that caused the error.

Requirements:
- Return exactly one short, specific sentence.
- Identify the failed operation, relation, constraint, evidence use or inference step.
- Use semantic roles instead of instance-specific names, dates, values, or positions.
- Do not add background, advice, headings, bullets, or generic evaluation.
- Do not state only the outcome, such as "incorrect answer" or "reasoning failure".
- Base the diagnosis only on the supplied trajectory and evaluator evidence.
- Failure may be caused by output that violates the task instructions, such as an unexpected format.

Task context:
{task_context}

Task instructions:
{method_specific_initial_instructions}

Message history:
{history_messages}

Ground truth code:
{ground_truth_code}

Test report:
{test_report}

Feedback:
{feedback}

Return only JSON with exactly one field:
{"failure_mechanism": "One short sentence describing the reusable failure mechanism."}
\end{lstlisting}
\subcaption{Prompt for AppWorld.}
\label{fig:prompt_b}
\endgroup

\caption{Failure mode extraction prompt for different tasks.}
\label{fig:failure_mode_prompt}
\endgroup

\begin{figure}[ht]
    \centering
    \includegraphics[width=0.5\linewidth]
        {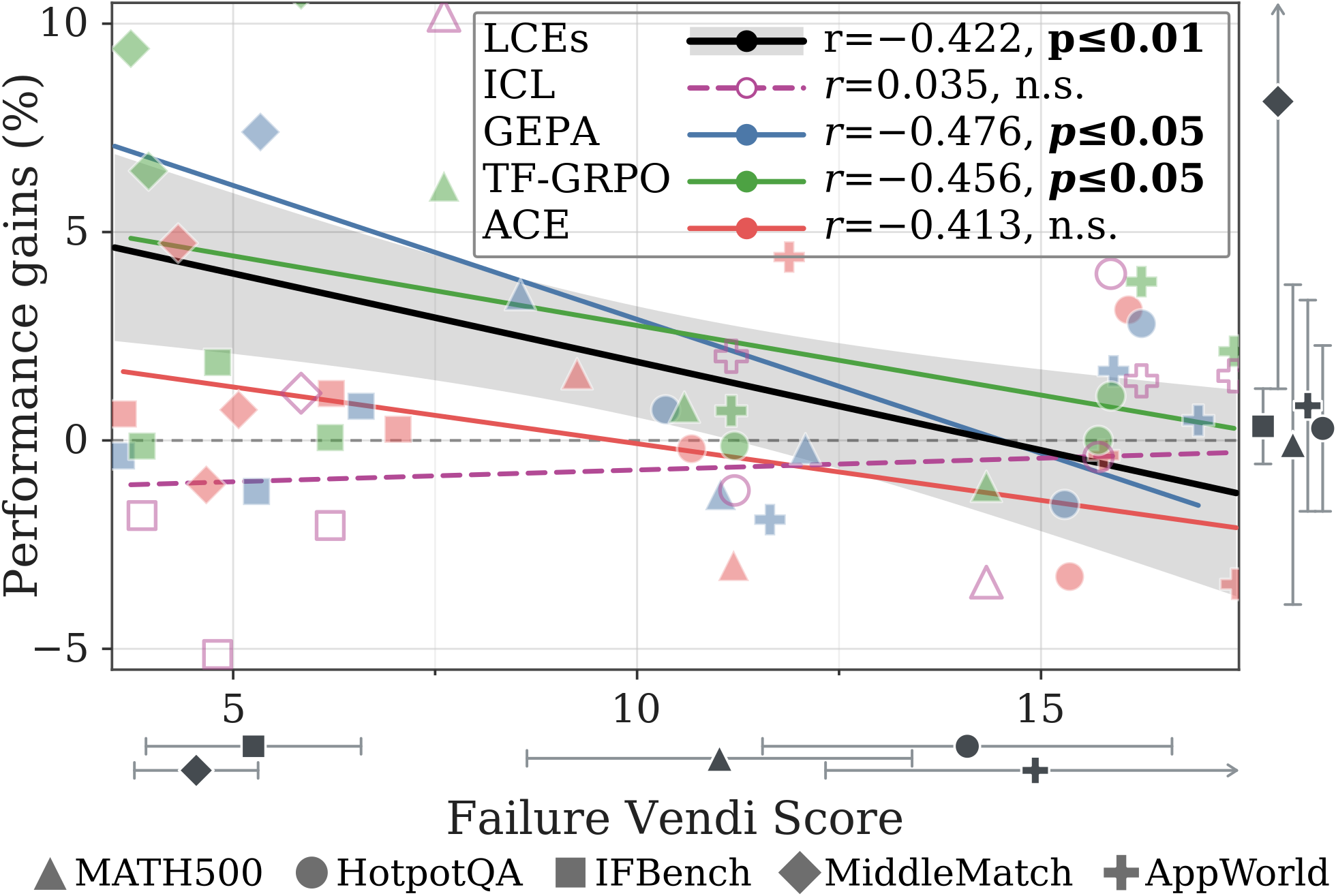}
    \caption{
    Correlation between LCE performance gains
    and the \emph{Vendi Score of failure modes} that are labeled by generator models themselves. 
    Markers and error bars show means and standard deviations on both axes for LCEs.
    The significant negative correlation indicates that
    LCEs tend to achieve larger performance gains when
    failure modes are less diverse.}
    % \jr{need to add Appworld after finishing training}}
    \label{fig:model_same_p_b}
    \label{fig:model_same_p}
    \vspace{-0.2in}
\end{figure}

Since LCEs often rely on the generator model for self-reflection, we also use each generator model to label the failure modes in its own rollouts. As shown in~\cref{fig:model_same_p}, the statistically significant negative correlation between Vendi Score and performance gains further suggests that LCEs construct more effective contexts when failure modes are less diverse.

\label{appendix:Qwen38}
% 第二张：Qwen3.8-27B Labeling，独立图片

\subsection{Variance Decomposition of Vendi Scores}
\label{appendix:calculation}
% \junyuan{Move this paragraph to appendix and add are ref above.}
We denote the Vendi Score for dataset $i$, model $j$, and LCE method $k$ by $v_{ijk}$. 
Let $\mathcal{Q}=\{\mathrm{D},\mathrm{M},\mathrm{L}\}$ denote the set of three factors, with $n_{\mathrm{D}}$, $n_{\mathrm{M}}$, and $n_{\mathrm{L}}$ denoting the numbers of datasets, models, and LCE methods, respectively. The total number of experimental settings is $n_{\mathrm{DML}}=n_{\mathrm{D}}n_{\mathrm{M}}n_{\mathrm{L}}$.
For each factor $q\in\mathcal{Q}$, let $\mu_a^q$ denote the mean Vendi Score for a specific element in $q$ indexed by $a$, averaged over the other two factors.
With the overall mean $\mu=\sum_{i,j,k}v_{ijk}/n_{\mathrm{DML}}$, the contribution of each main effect to the total variance is
$
\sigma_q^2=\sum_{a=1}^{n_q}\left(\mu_a^q-\mu\right)^2/n_q,\quad q\in\mathcal{Q}.
$
The residual is
$
\sigma_{\mathrm{res}}^2=\sum_{i,j,k}\left(v_{ijk}-\mu_i^{\mathrm{D}}-\mu_j^{\mathrm{M}}-\mu_k^{\mathrm{L}}+2\mu\right)^2/n_{\mathrm{DML}},
$
which captures interactions among the factors. These components sum to the total variance:
$
\sigma_{\mathrm{total}}^2
% =\sum_{i,j,k}(v_{ijk}-\mu)^2/n_{\mathrm{DML}}
=\sum_{q\in\mathcal{Q}}\sigma_q^2+\sigma_{\mathrm{res}}^2.
$ 
We report the variance for each factor and residual as a proportion of the total variance, $\sigma_q^2/\sigma_{\mathrm{total}}^2$, where $q\in\mathcal{Q}\cup \{\mathrm{res}\}$. Here, a larger variance proportion indicates greater differences in failure-mode diversity across the levels of the corresponding factor.

\subsection{Experimental Details for Case Studies}
\label{appendix:case_study}

When calculating the false failure rate, we directly measure the proportion of predictions that contain the ground-truth answer in the final response but are marked as incorrect.
We remove some refusal responses, e.g., cases where the model cannot find the correct answer but repeats text containing the ground-truth answer, from false failure predictions by matching phrases such as \emph{unable to determine} to avoid additional noise.
For calculating the format compliance rate on MATH500, we measure the proportion of final answers from which our Qwen-Verify can successfully identify the \texttt{\textbackslash boxed\{\}}.

\section{\text{Pass@K} and \text{Pass\textasciicircum K} Evaluation}
\label{appendix:passk}
\begin{figure*}[ht]
    \centering
    \setlength{\tabcolsep}{1pt}
    \renewcommand{\arraystretch}{0}
    \includegraphics[width=\textwidth]{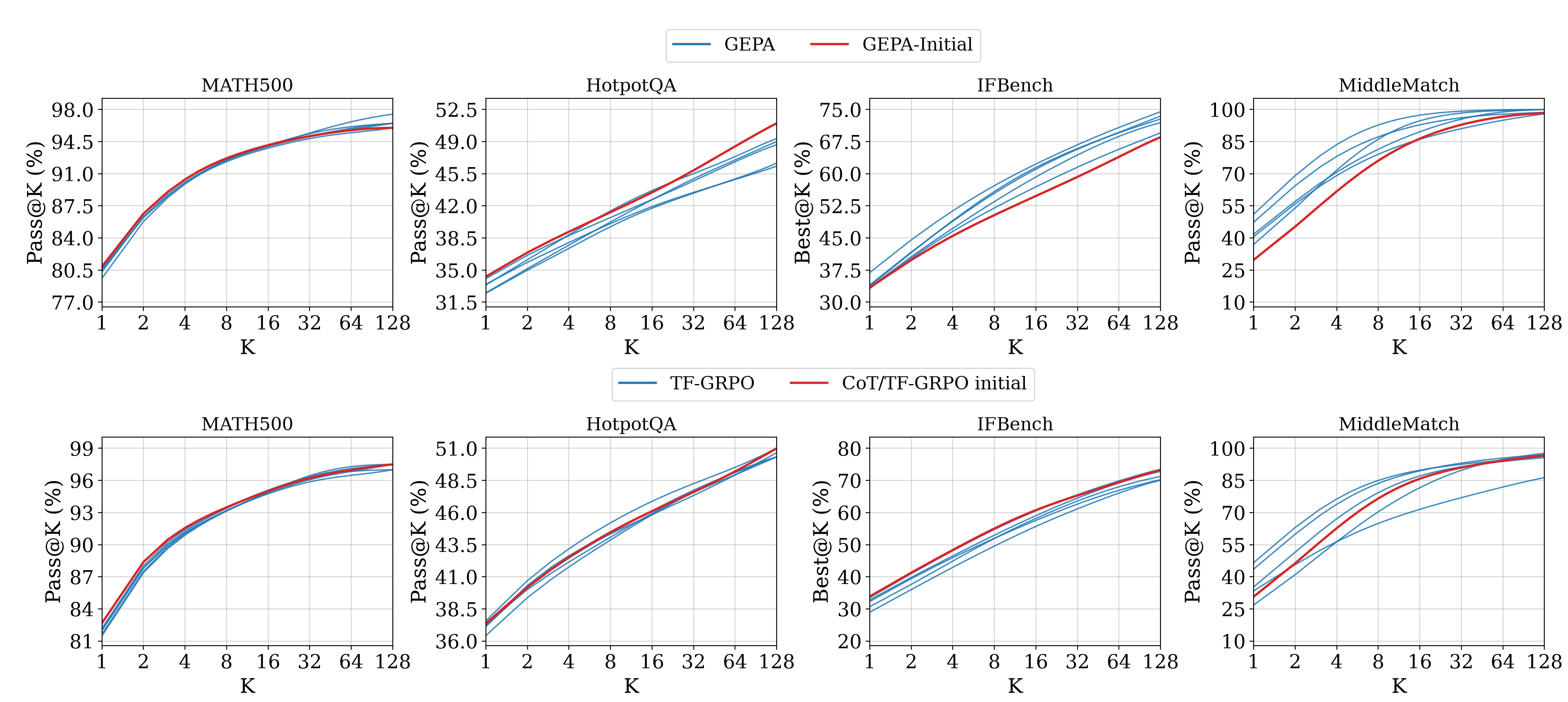}
    \caption{Pass@K or Best@K on four benchmarks. \textbf{Top row}: GEPA. \textbf{Bottom row}: TF-GRPO. Each figure compares the different optimized contexts by GEPA/TF-GRPO versus their initial performance.}
    \vspace{-0.1in}
    \label{fig:pass_at_k}
\end{figure*}

\textbf{LCEs Can Introduce New (But Possibly Bad) Knowledge.} It was debatable whether the (none) improvement at Pass@1 actually reveals the gain of new knowledge/capability or just better exploits existing knowledge/capability~\citep{PassAtK}.
Obviously, the former is a more valuable gain than simply reshaping the output distribution, and is expected with a lot of data.
% Due to the limited sample size and the large variance of LLM generation, it is debatable if the difference caused by the learned context is too small to observe.
% When an LLM preserves a piece of knowledge, the knowledge may be elicited only after enough trials~\citep{LetsVerifyStepByStep}.
To uncover the type of gain, we report Pass@K or Best@K (on IFBench) for $\text{K} \in [1, 128]$ in \cref{fig:pass_at_k}. Pass@K is the fraction of problems solved by at least one of K generated samples and Best@K represents the expected maximum score across K generated samples. All Pass@K or Best@K results are computed from the same set of 128 generated samples per problem using the unbiased estimator 
% \junyuan{Wrap the formula in line. This is never referred. Do not use numbered eq.} \jr{done}
$
\text{Best@}\text{K} \;=\; \mathbb{E}_{\text{problems}}\!\left[\, \sum_{i=\text{K}}^{128} \left({\binom{i-1}{\text{K}-1}}\Big/{\binom{128}{\text{K}}}\right) s_{(i)} \,\right],
$
where $s_{(i)}$ is the $i$-th smallest score among the 128 generated samples. Pass@K is the special case of Best@K with binary scores $s_i \in \{0, 1\}$.
The comparison between Pass@K and Pass@1 can indicate how the context impacts the performance.
% we leverage the Pass@K (or best@k) curve to reveal if new knowledge is introduced.
(1) For performance gain at Pass@1, if the gain is maintained with larger K, we say there is new knowledge introduced. Otherwise, the learning is less valuable.
(2) For performance loss at Pass@1, it is also worth examining the curve. If misinformation is introduced, the curve of learned context should be globally lower than the baseline.

\begin{figure*}[ht]
    \centering
    \includegraphics[width=\textwidth]{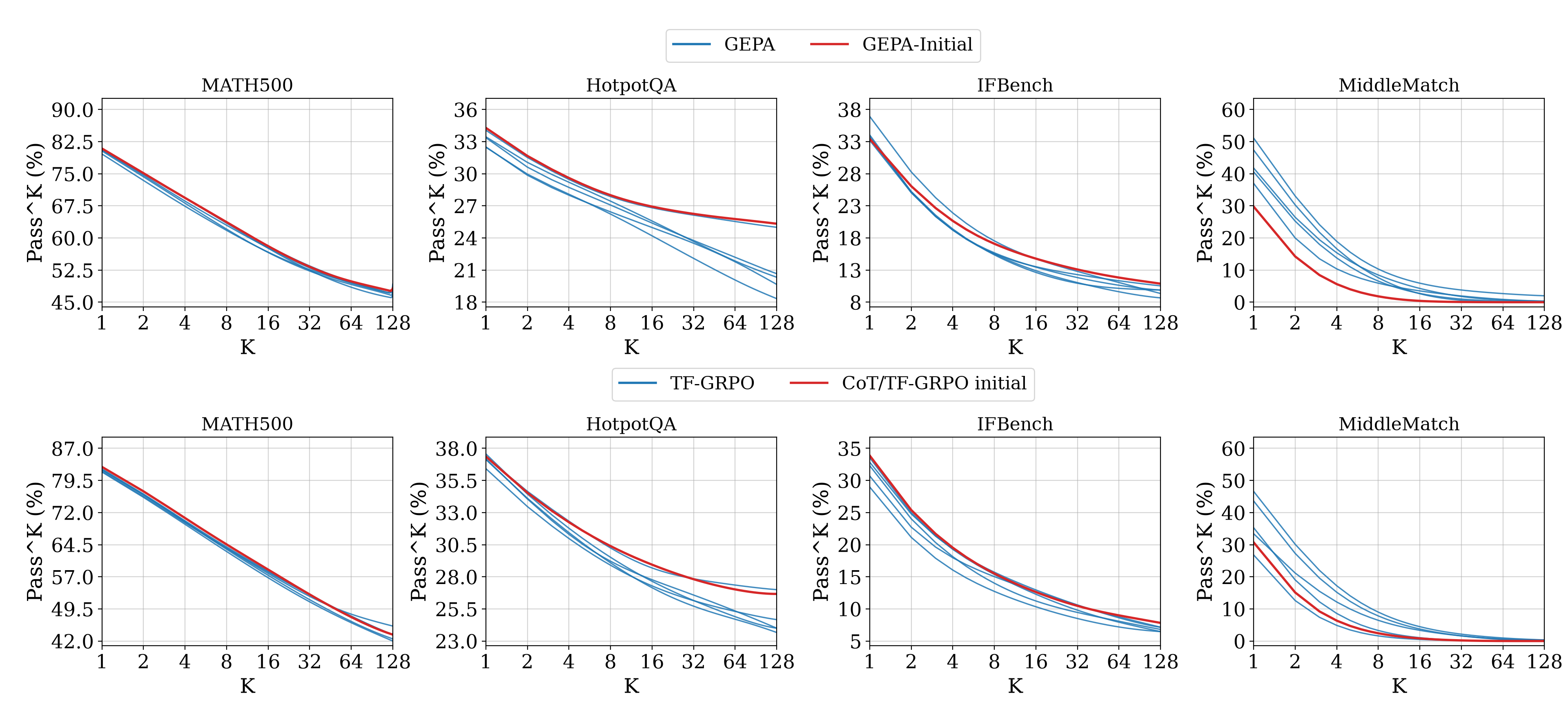}
    \caption{\text{Pass\textasciicircum K} Evaluation on four benchmarks. \textbf{Top row}: GEPA. \textbf{Bottom row}: TF-GRPO. Each figure compares the different optimized contexts by GEPA/TF-GRPO versus their initial performance.
    % \jr{this picture will be changed}
    }
    \label{fig:pass_k_times}
    % \vspace{-0.2in}
\end{figure*}

In \cref{fig:pass_at_k}, we observe that LCEs can introduce both beneficial and harmful knowledge. On HotpotQA, GEPA's context underperforms its initial across all attempts, with the gap persisting up to Pass@128, indicating that LCEs can inject harmful knowledge that significantly biases the model and is difficult to repair by additional sampling. On MiddleMatch, both TF-GRPO and GEPA improve over their initial performance from Pass@1 to Pass@K in most attempts. Moreover, on IFBench and MATH500, some of GEPA's contexts match or slightly fall below their initials at Pass@1 but exceed them at larger K. These results suggest that the LCEs can also introduce useful knowledge hinges in the data and trial. 
Besides Pass@K, we also report \text{Pass\textasciicircum K}, which evaluates the model's ability to consistently produce correct answers across \(K\) attempts. Given 128 sampled responses for each problem, \text{Pass\textasciicircum K} is computed as
$
\text{Pass}\text{\textasciicircum}\text{K}
\;=\;
\mathbb{E}_{\text{problems}}
\left[
\binom{c}{\text{K}} / \binom{128}{\text{K}}
\right],
$
where $c$ is the number of correct answers in the 128 attempts. For IFBench, we compute the expected product of the scores of K attempts as the \text{Pass\textasciicircum K}. As shown in \cref{fig:pass_k_times}, GEPA and TF-GRPO fail to achieve their initial performance from \text{Pass}\text{\textasciicircum}\text{1} to \text{Pass}\text{\textasciicircum}\text{K}, indicating that the harmful knowledge will undermine the reliability of LLMs to generate correct answers.
% We therefore provide the detailed results in Appendix~\ref{sec:pass_k_times}.

\label{sec:pass_k_times}

% \jr{this need to be revised}

\section{Additional Observations in Scaling Experiments}
\label{appendix:scaling_law}
\textbf{Good Context Boosts Smaller Model to Match Larger Ones.}
On MiddleMatch, we observe a more consistent trend than other benchmarks:
(1) Both methods with a 32B Reflector boost the performance of the 8B LLM, and GEPA even pushes the 8B LLM (42.33 $\to$ 60.40) to match the 32B CoT baseline (60.67), though with larger variance (12.93 on 8B vs.\ 3.32 on 32B).
(2) Both methods with an 8B Reflector enhance the 32B LLM compared to their own initial prompts, while only TF-GRPO with an 8B Reflector outperforms the best CoT baseline.

\par\noindent
\textbf{Smaller Model can unlock the capability of Larger Model.}
On IFBench, GEPA generally benefits from scaling up the reflector (\cref{fig:boxplot_results}), whereas TF-GRPO under the 32B generator slightly underperforms. Remarkably, while the 8B and 32B models show similar CoT performance (32.35 vs.\ 33.47), using an 8B reflector alone is sufficient to unlock the 32B model's potential: GEPA pushes the 32B model to 39.08.

\section{Case Study on TFGRPO with more variants}
\label{appendix:tfgrpo_case_study}
We add two inference variants that use a fixed number of experiences or in-context examples:
(1) All-Exps RAG retrieves experiences from the full set of extractor-proposed experiences accumulated during TF-GRPO training.
% (2) All-Exps Random randomly samples experiences from all experiences produced by TF-GRPO (Final-Exps).
% same pool used by All-Exps RAG.
(2) Final-Exps RAG retrieves experiences from the final experience pool produced by TF-GRPO (Final-Exps).
The full results are shown in~\cref{tab:grpo-ablation-full}.

\begin{table*}[t]
    \caption{Performance under different experience utilization and retrieval
    settings. \textbf{R}, \textbf{E}, and $\boldsymbol{\phi}$ denote training
    rollouts, knowledge extraction, and experience consolidation, respectively.
    $k$ denotes the number of examples or experiences used, with \textcolor{gray}{var.} indicating that the number varies depending on the optimized experiences. \textbf{Bold} and \underline{underlined} values indicate the best and second-best results.
    Q, G, and M denote Qwen3, Gemma4, and Min3, respectively. 
    % \junyuan{I removed some rows. Provide the full table in appendix.} \jr{done_}
    }
    \label{tab:grpo-ablation-full}
    \vspace{-0.05in}
    \footnotesize
    \centering
    \setlength{\tabcolsep}{3pt}
    \renewcommand{\arraystretch}{1.1}
    \begin{tabular}{@{}lcccc|cccc@{}}
    \toprule

    \multirow{2}{*}{\textbf{Method}}
    & \multirow{2}{*}{\textbf{R}}
    & \multirow{2}{*}{\textbf{E}}
    & \multirow{2}{*}{$\boldsymbol{\phi}$}
    & \multirow{2}{*}{$k$}
    & \textbf{MiddleMatch}$^{\mathrm{Q}}$
    & \textbf{MATH500}$^{\mathrm{G}}$
    & \textbf{MATH500}$^{\mathrm{M}}$
    & \textbf{IFBench}$^{\mathrm{G}}$ \\

    \cmidrule(lr){6-7}
    \cmidrule(lr){8-9}

    & & & &
    & \multicolumn{2}{c}{$\mathrm{ICL}>\mathrm{LCE}>\mathrm{CoT}$}
    & \multicolumn{2}{c}{$\mathrm{LCE}\geq\mathrm{CoT}>\mathrm{ICL}$} \\

    \midrule

    CoT
    & & & & 0
    & $31.27{\scriptscriptstyle\pm2.10}$
    & $67.70{\scriptscriptstyle\pm1.04}$
    & $87.50{\scriptscriptstyle\pm1.06}$
    & $39.76{\scriptscriptstyle\pm2.22}$ \\

    ICL
    & \checkmark & & & 3
    & $\mathbf{46.07}{\scriptscriptstyle\pm4.80}$
    & $\mathbf{77.90}{\scriptscriptstyle\pm3.54}$
    & $68.10{\scriptscriptstyle\pm2.41}$
    & $34.63{\scriptscriptstyle\pm2.17}$ \\

    \midrule

    GRPO-ICL RAG
    & \checkmark & & & 3
    & $\underline{44.00}{\scriptscriptstyle\pm3.46}$
    & $\underline{74.10}{\scriptscriptstyle\pm2.25}$
    & $26.50{\scriptscriptstyle\pm4.23}$
    & $37.41{\scriptscriptstyle\pm1.44}$ \\

    All-Exps Random
    & \checkmark & \checkmark & & 3
    & $32.00{\scriptscriptstyle\pm6.00}$
    & $69.70{\scriptscriptstyle\pm0.91}$
    & $87.40{\scriptscriptstyle\pm0.55}$
    & $\underline{40.92}{\scriptscriptstyle\pm2.07}$ \\

    All-Exps RAG
    & \checkmark & \checkmark & & 3
    & $39.07{\scriptscriptstyle\pm11.87}$
    & $68.30{\scriptscriptstyle\pm2.89}$
    & $\mathbf{88.50}{\scriptscriptstyle\pm1.27}$
    & $38.78{\scriptscriptstyle\pm1.67}$ \\

    % \midrule

    Final-Exps
    & \checkmark & \checkmark & \checkmark & \textcolor{gray}{var.}
    & $37.73{\scriptscriptstyle\pm8.57}$
    & $73.80{\scriptscriptstyle\pm4.12}$
    & $\underline{88.30}{\scriptscriptstyle\pm1.15}$
    & $\mathbf{41.63}{\scriptscriptstyle\pm1.21}$ \\

    Final-Exps RAG
    & \checkmark & \checkmark & \checkmark & 3
    & $38.53{\scriptscriptstyle\pm5.62}$
    & $68.80{\scriptscriptstyle\pm1.04}$
    & $87.90{\scriptscriptstyle\pm1.88}$
    & $40.00{\scriptscriptstyle\pm2.11}$ \\

    \bottomrule
    \end{tabular}
    \vspace{-0.05in}
\end{table*}

% \subsection{Context Length vs. Content Quality}
\section{Which Is Better Context: More High-Quality Experiences Or Fewer Experiences?}
\label{appendix:length}

\textbf{In-Context Demonstrations Can Mislead LLMs Into Making Errors.}
In settings where LCE $\geq$ CoT $>$ ICL, replacing random demonstrations with retrieved training demonstrations reduces accuracy sharply from 68.10\% to 26.50\%. Our analysis identifies two main causes: (1) lengthy context impairs reasoning, and (2) demonstrations can distract models from the target problem.
Training demonstrations are selected from problems with both successful and failed attempts. These problems are challenging for the model and may require more extensive reasoning to distinguish correct solutions from plausible alternatives, making GRPO-ICL demonstrations substantially longer than ICL demonstrations. To examine the effect of context length, we reduce the number of retrieved demonstrations to match the total context length of ICL, as shown in~\cref{tab:icl_prompt_length}. This restores accuracy to approximately 68.10\%, suggesting that excessive context length may contribute to the performance degradation, potentially by making it harder for the model to focus on critical reasoning steps.
Meanwhile, we observe cases where Min3 answers a demonstration question instead of the target problem. Thus, we report the ratio of responses that answer an in-context demonstration rather than the target problem. As shown in \cref{tab:answer_mistakes}, we find that Min3 is difficult to distinguish demonstrations from the target problem, leading to these failures. LCEs may mitigate this issue by extracting knowledge from demonstrations, leaving only target question to answer in the context.

\begin{table}[ht]
  \centering
  \small
  \vspace{-0.1in}
  \caption{Performance under different ICL settings with varying prompt lengths.}
  \label{tab:icl_prompt_length}
  \begin{tabular}{lc | cc}
    \toprule
    \textbf{Method} & $k$ & \textbf{Avg. Prompt Tokens} & \textbf{Accuracy} \\
    \midrule
    ICL &3       & 3,859  & 68.10 \\
    GRPO-ICL RAG &1 & 3,807  & 68.50 \\
    GRPO-ICL RAG &3 & 11,112 & 26.50 \\
    \bottomrule
  \end{tabular}
  \vspace{0in}
\end{table}

\begin{table}[ht]
  \centering
  \small
  \vspace{-0.1in}
  \setlength{\tabcolsep}{4pt}
  \caption{Ratio of the predictions that incorrectly match the in-context demonstration answer.}
  \label{tab:answer_mistakes}
  \begin{tabular}{l c| c} % Original: {l c| cc}
    \toprule
    \textbf{Method} & $k$
    & \textbf{Answer Any ICL}
    % & \textbf{Answer Final ICL}
    \\
    \midrule
    ICL          & 3 & 15.10\%
    % & 13.50\%
    \\
    GRPO-ICL RAG & 1 & 14.90\%
    % & 14.90\%
    \\
    GRPO-ICL RAG & 3 & 64.30\%
    % & 59.40\%
    \\
    \bottomrule
  \end{tabular}
  \vspace{0in}
\end{table}

\par
\begin{figure}[h]
    \centering
    \includegraphics[width=0.6\linewidth]{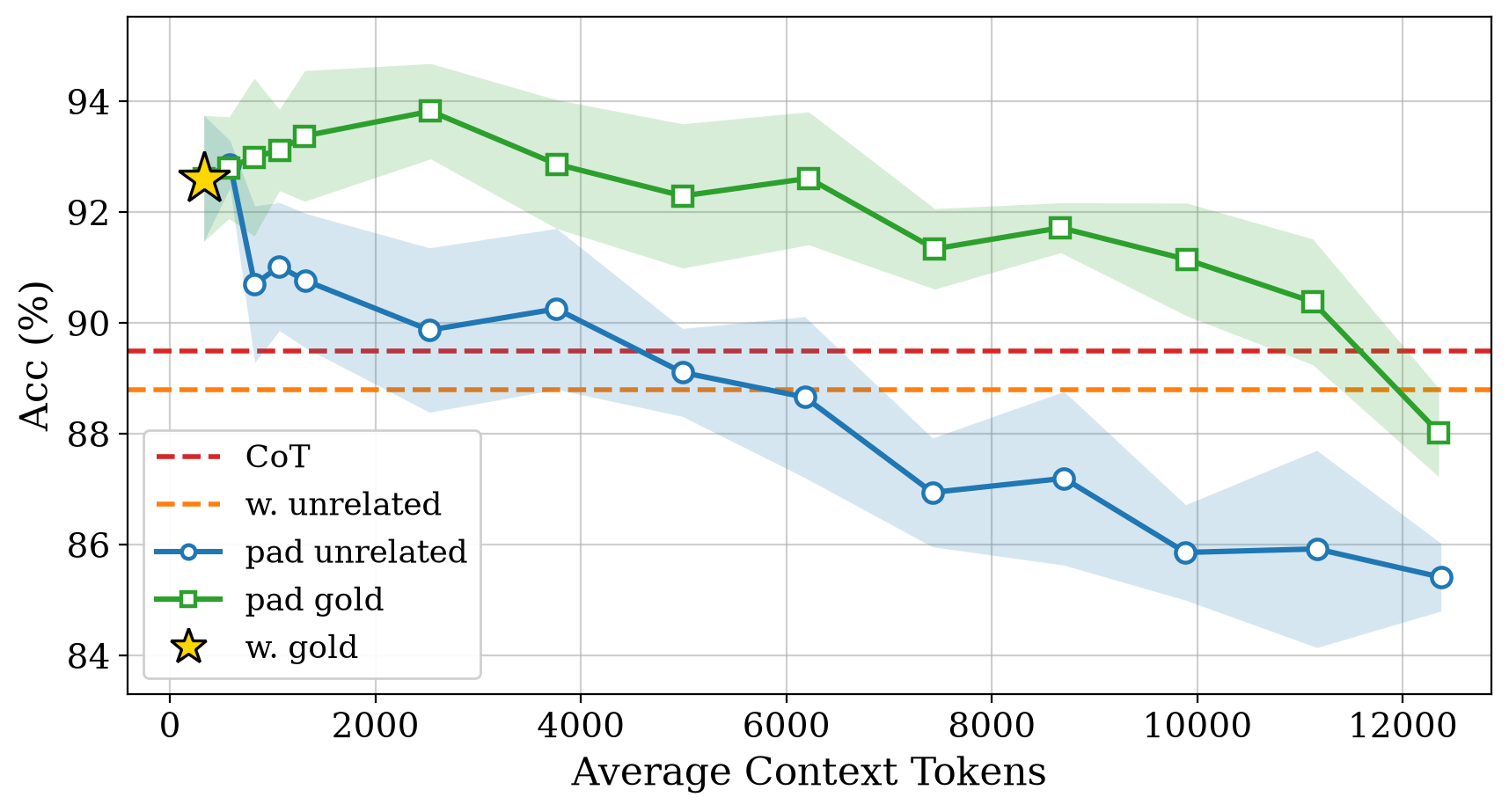}
    \caption{Ablation on context length. As high-quality context grows longer, performance eventually drops below both baselines. }
    % \junyuan{The unrelated-ICL experiment should not include any golden examples.}}
    % \junyuan{but unrelated-ICL has failed using short context, I want to say: (1)high-quality but lengthy is bad (2)high-quality context cannot be used when too many unrelated context}}
    \label{fig:length}
\end{figure}

\noindent
\textbf{Controlled Study With Gold ICL.}
% Although the above results suggest that lengthy context can be harmful, the performance recovery may stem from removing low-quality dict-format experiences rather than reducing context length. 
% Although the above results suggest that lengthy context may be harmful,
The above results suggest that lengthy contexts may be harmful, which is consistent with previous work. \rev{For example, \citet{zhang-etal-2025-always} observed that performance improves with a few-shot ICL setup but declines with many-shot ICL, while \citet{contextrot} showed that lengthy contexts can introduce noisy information and consequently hurt performance.}
However, another possible explanation for the failures is that GRPO-ICL demonstrations are of lower quality.
To separate the effects of context length and quality, we conduct an additional controlled study using ICL.
Since directly evaluating the ``quality'' of an evolving prompt is difficult and inherently subjective, we therefore construct MATH-Sync by paraphrasing each MATH500 question conditioned on its ground-truth reasoning trace. We validate correctness via majority voting and remove duplicates (cosine similarity $>0.9$ or identical final answers), yielding 318 (synthetic, original) pairs. For each test problem, we use its paired synthetic paraphrase as a \emph{gold ICL demonstration}, which shares the underlying solution process with the test problem by construction and thus provides a controlled high-quality context.
% Therefore, we construct an additional training set, MATH-Sync, by paraphrasing each MATH500 question conditioned on its ground-truth reasoning trace. We validate correctness via majority voting and discard duplicates (cosine similarity > 0.9 or identical final answer), leaving 318 (synthetic, original) pairs.
% We therefore use MATH-Sync to construct a controlled experiment in which context quality is held at a known upper bound. For each test problem, we treat the synthetic paraphrase from its data pair as a \emph{gold ICL demonstration}, since the paraphrase shares the underlying solution process with the test problem by construction.
We compare three baselines: (i) the original CoT prompt with no demonstrations, (ii) ICL with irrelevant demonstrations sampled from unrelated problems, and (iii) gold ICL using the matched synthetic paraphrase. As shown in ~\cref{fig:length}, gold ICL substantially outperforms both other baselines, confirming that high-quality demonstrations are genuinely valuable when they are concise.

\label{sec:benchmark-context-length}
% \junyuan{To the question in the title, the experiment design is not a clear correspondence. Why you include unrelated experiences instead of random experiences from training set?}
% \junyuan{Need to re-think the hypothesis in this section. This seems focus on ICL. I feel it does not quite fit into the story flow here.}

\par\noindent
\textbf{Long Gold Context Is Worse Than Concise Context.}
We then progressively pad the gold ICL prompt in two ways: \textbf{(a)} by repeating the gold demonstration itself, holding quality constant while increasing length; and \textbf{(b)} by inserting additional ICL demonstrations sampled from unrelated problems, increasing length while diluting quality. In the early regime, both padding strategies further improve performance, consistent with the intuition that ``more demonstrations help.'' However, as the prompt continues to grow, both curves eventually fall below all three baselines, including the original CoT prompt with no demonstrations at all. Crucially, this degradation occurs even under the repeated-gold setting, where every additional token comes from a high-quality, perfectly aligned demonstration. Length alone is therefore sufficient to drive performance below baseline. Together with the gold ICL result, this shows that the value of an injected experience is not monotonic in either quality or quantity: high-quality information improves performance up to a point, after which length-induced degradation overwhelms the benefit.

\section{Mixed-effects Analysis on Failure-mode Diversity}
\label{app:mixed_effects}
\rev{
To further control for the potential influence of other factors, including dataset identity, model identity, LCE methods, and initial performance, we use a linear mixed-effects model to test our hypothesis that LCEs can achieve larger performance gains when failure modes are less diverse. 
Specifically, performance gain is defined as the outcome variable, incorporating failure-mode diversity, dataset identity, and initial performance as fixed effects, while treating models and LCE methods as random effects. 
We report the $\beta$ and corresponding p-value from the t-test.
As shown in~\cref{tab:mixed_effects}, the coefficient of failure-mode diversity remains significantly negative ($\beta = -0.793, p\leq0.05$), which is consistent with our finding that LCEs can be more effective when addressing more homogeneous failure modes. Moreover, the coefficient for initial performance and the individual coefficients for the four datasets relative to AppWorld are not significant. In conclusion, the differences in the effectiveness of LCEs are associated with the failure-mode diversity.
}
% \begin{figure}[ht]
%     \centering
%     \includegraphics[width=0.5\linewidth]
%         {figs/app/mixed_effect7.png}
%     \caption{
%     Correlation between LCE performance gains
%     and the \emph{Failure Vendi Score} using a linear mixed-effects model. 
%     Markers and error bars show means and standard deviations on both axes for LCEs.
%     The significant correlation implies that the LCEs can improve the context better when facing less heterogeneous failure modes.}
%     \label{fig:mixed_effects}
%     % \vspace{-0.2in}
% \end{figure}

\begin{table}[t]
    \centering
    \caption{Coefficients and corresponding p-values for each variable in the linear mixed-effects model, with performance gain as the outcome variable. We observe that the coefficient of failure-model diversity is significantly negative, implying that the LCEs can improve the context better when facing less heterogeneous failure modes.}
    \label{tab:mixed_effects}
    \begin{tabular}{l|cc}
        \toprule
        Variable & $\beta$ & p-value \\
        \midrule
        Failure-mode diversity & $-0.793$ & $\boldsymbol{p \leq 0.05}$ \\
        Initial performance    & $-0.038$ & n.s. \\
        % \midrule
        HotpotQA       & $-2.070$ & n.s. \\
        IFBench        & $-5.427$ & n.s. \\
        MATH500        & $-1.698$ & n.s. \\
        MiddleMatch    & $+2.658$ & n.s. \\
        \bottomrule
    \end{tabular}
\end{table}

\section{Additional Analysis of Other Factors Potentially Associated with Performance Gains}
\label{app:nonsig}
\subsection{Analysis on optimization budget}
\begin{figure}[ht]
    \centering
    \includegraphics[width=0.48\linewidth]{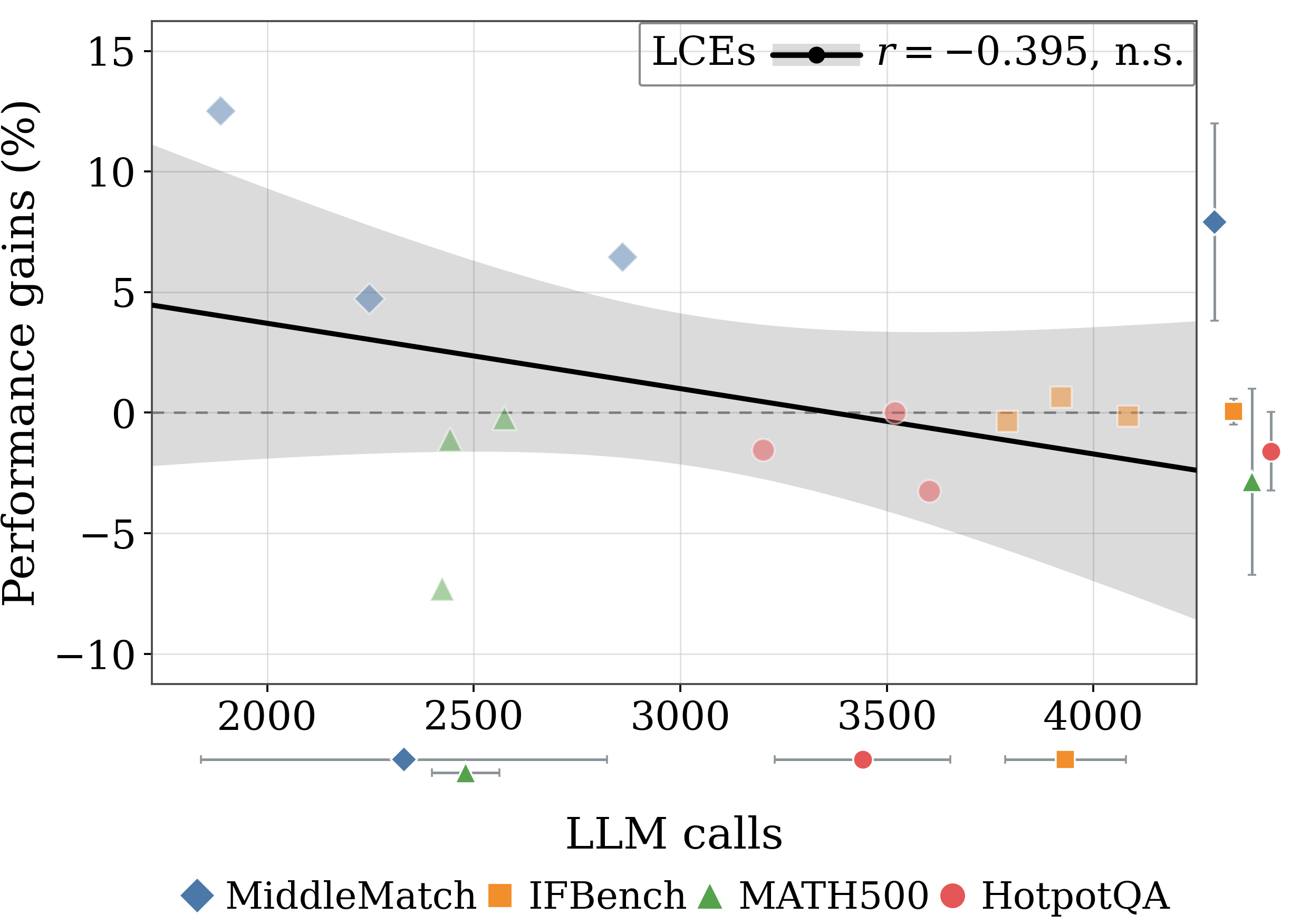}
    \includegraphics[width=0.48\linewidth]{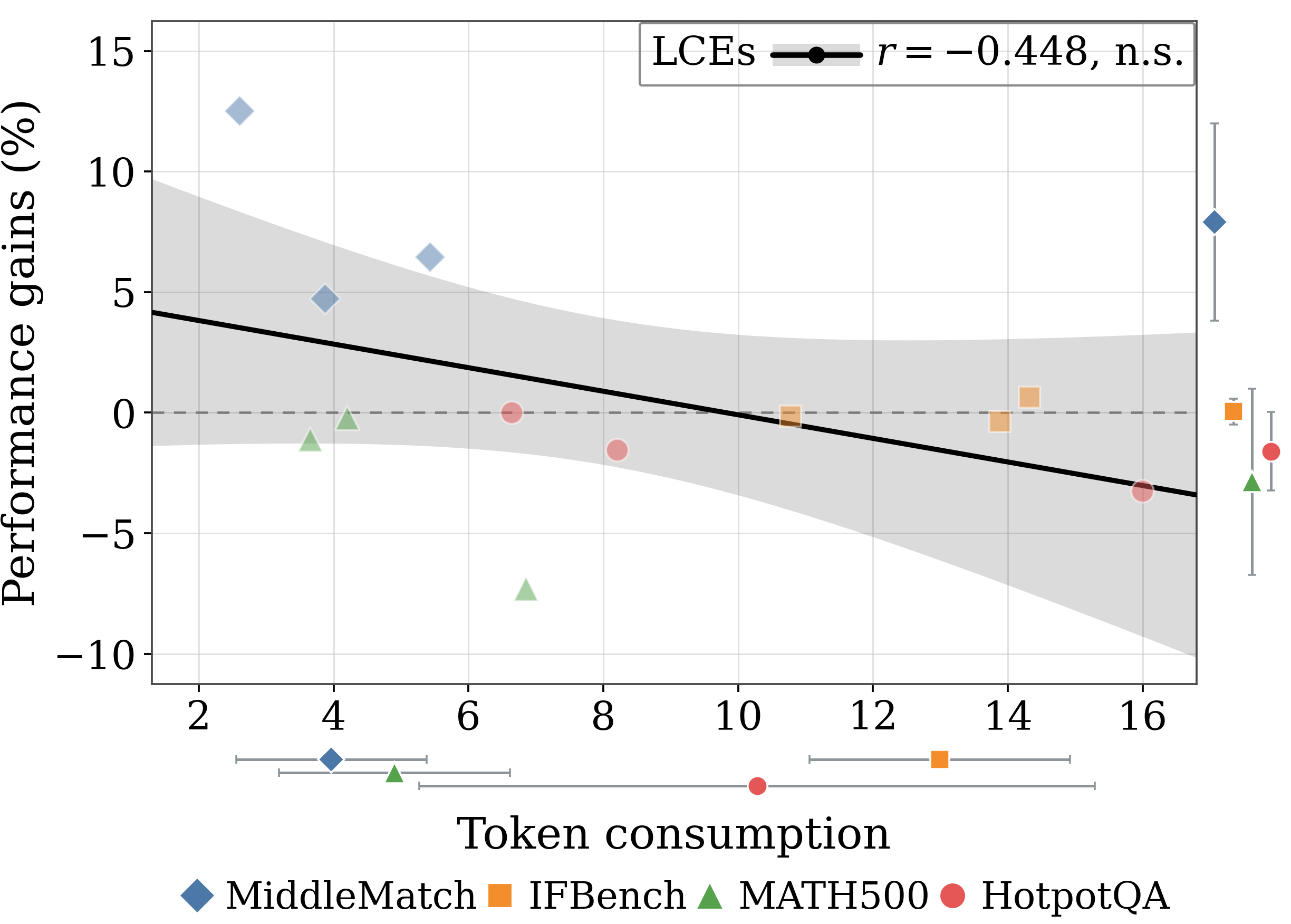}
    \vspace{-0.05in}
    \caption{
    Correlation between LCE performance gains and optimization budgets (\textbf{Left:} LLM calls. \textbf{Right:} token consumption in millions).
    Markers and error bars represent means and standard deviations, respectively, on both axes.
    Neither correlation is significant. We find no evidence that larger optimization budgets are associated with more LCE performance gains.
    }
    \label{fig:budget}
    \vspace{-0.05in}
\end{figure}

\rev{
Since different LCE designs inevitably lead to different optimization budgets during context engineering, we further investigate the relationship between optimization budgets and the performance gains achieved by LCEs. Specifically, we quantify the optimization budget using two metrics: the number of LLM calls during optimization and the corresponding token consumption, including both input and output tokens.
To assess whether different optimization budgets are associated with differences in LCE effectiveness, we conduct correlation analyses using the Qwen3 results across four datasets.
As shown in~\cref{fig:budget}, no significant correlations are observed between performance gains and optimization budgets. A potential explanation is that additional optimization budgets may be spent evaluating ineffective candidates or repeatedly refining similar instructions, leading to the result that more optimization budgets are insufficient to improve the context.}

\subsection{Analysis on other baseline}
\rev{We additionally append the instruction “Answer only what the question asks. Avoid extraneous details.” to both the initial prompts and the contexts optimized by each LCE method across models and datasets. 
We report the performance gains from the initial to the optimized contexts, with and without this prompt, in~\cref{tab:prompt_ablation}. We then perform a paired t-test comparing these gains across all evaluated settings and find that the non-significant result provides no evidence that simplified explanations can harm the LCEs' effectiveness.}

\begin{table*}[t]
\centering
\caption{Performance gains of LCEs with and without additional prompts across models on different tasks. 
Our experiments cannot provide sufficient evidence that simplified explanations can affect the LCEs' performance.}
\label{tab:prompt_ablation}
\scriptsize
\renewcommand{\arraystretch}{0.9}
\setlength{\tabcolsep}{4pt}
\resizebox{0.9\textwidth}{!}{%
\begin{tabular}{ll|cccccc}
\toprule
\multirow{2}{*}{\textbf{Model}}
& \multirow{2}{*}{\textbf{Dataset}}
& \multicolumn{3}{c}{\textbf{w/o Prompt}}
& \multicolumn{3}{c}{\textbf{w/ Prompt}} \\
\cmidrule(lr){3-5}
\cmidrule(lr){6-8}
& & GEPA & TF-GRPO & ACE & GEPA & TF-GRPO & ACE \\
\midrule
\multirow{4}{*}{Qwen3}
& MATH500     & -0.20 & -1.10 & -7.30 & -0.80 & +0.80 & -5.70 \\
& HotpotQA    & -1.54 & +0.00 & -3.26 & -0.33 & +1.27 & -2.07 \\
& IFBench     & -0.37 & -0.14 & +0.64 & -1.16 & -1.16 & +1.12 \\
& MiddleMatch & +12.53 & +6.47 & +4.73 & +12.93 & +6.60 & +5.40 \\
\midrule
\multirow{4}{*}{Gemma4}
& MATH500     & +3.50 & +6.10 & +1.60 & +2.20 & +9.00 & +1.10 \\
& HotpotQA    & +0.73 & -0.13 & -0.20 & -0.53 & -0.33 & -0.47 \\
& IFBench     & -1.22 & +1.87 & +1.13 & -1.94 & +0.54 & +1.77 \\
& MiddleMatch & +7.40 & +10.80 & +0.73 & +5.27 & +3.60 & +3.40 \\
\midrule
\multirow{4}{*}{Min3}
& MATH500     & -1.30 & +0.80 & -3.00 & +0.60 & -1.60 & -6.90 \\
& HotpotQA    & +2.80 & +1.07 & +3.13 & -0.27 & +0.47 & +0.93 \\
& IFBench     & +0.82 & +0.07 & +0.27 & +2.55 & -0.20 & -0.03 \\
& MiddleMatch & +22.20 & +9.40 & -1.07 & +14.80 & +0.33 & +5.13 \\
\bottomrule
\end{tabular}%
}
\end{table*}

\subsection{Analysis on baseline task accuracy}
\begin{figure}[ht]
    \centering
    \includegraphics[width=0.48\linewidth]
        {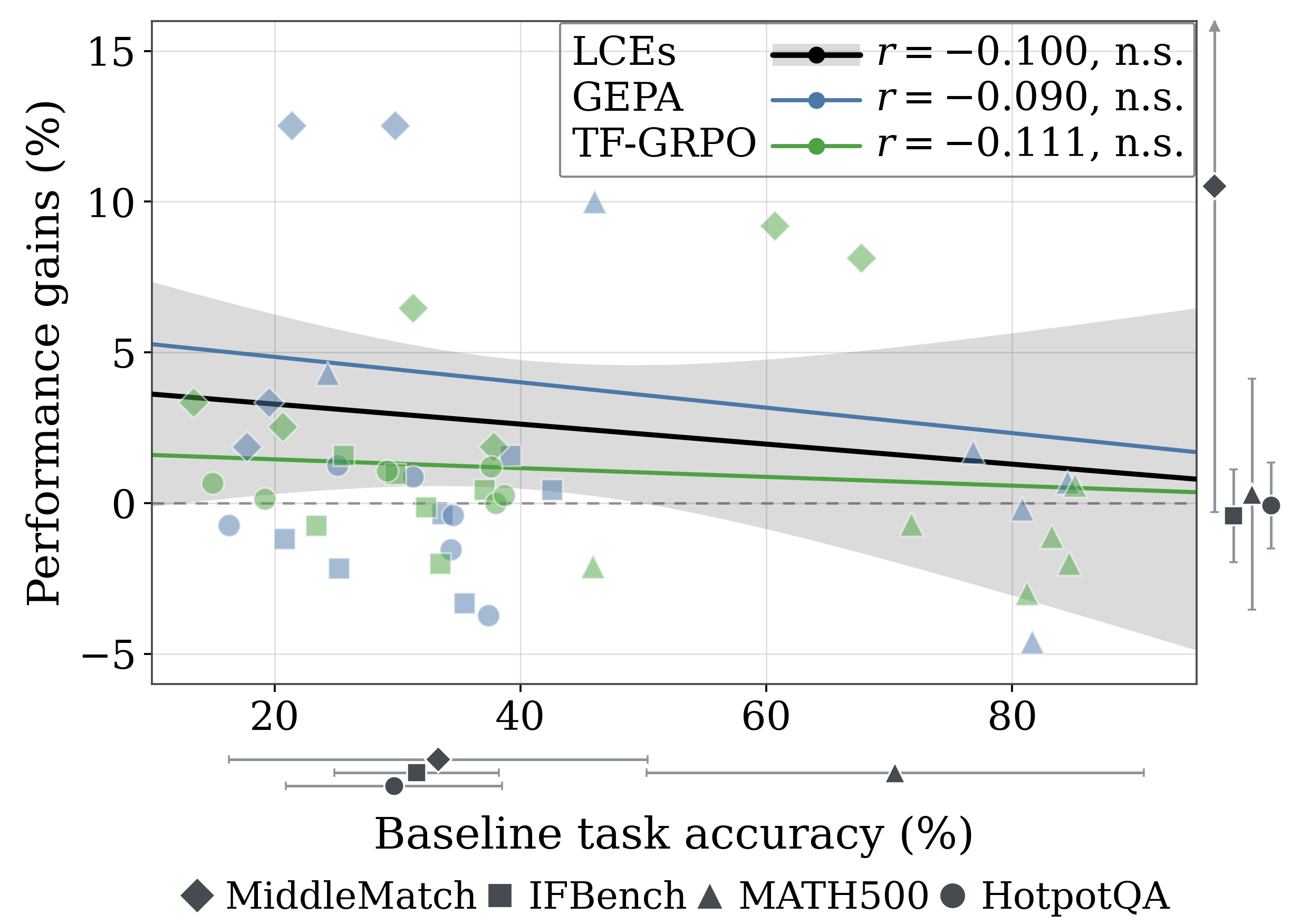}
    \includegraphics[width=0.48\linewidth]
        {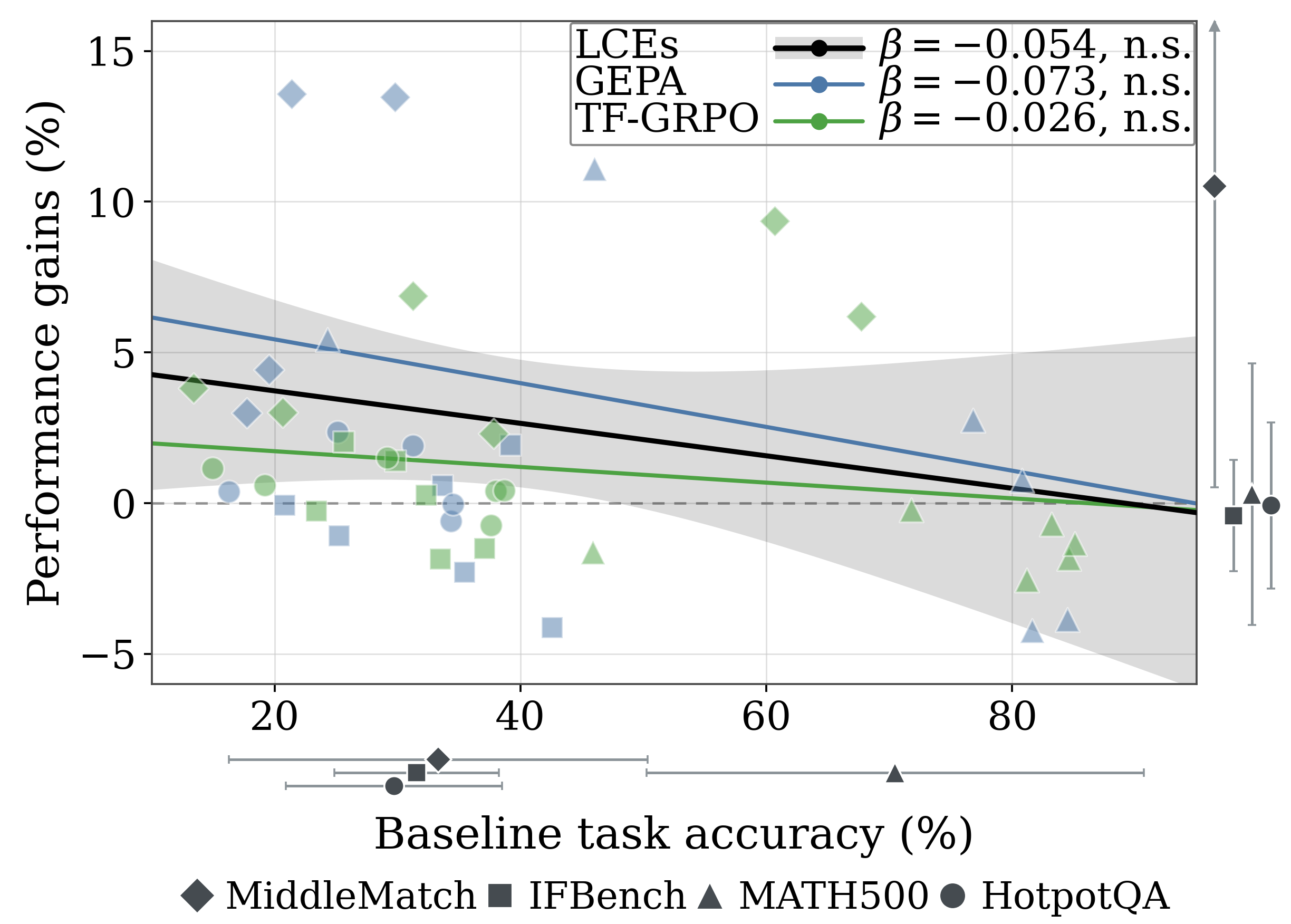}
    \vspace{-0.05in}
    \caption{
    Correlation between LCE performance gains
    and the \textbf{Left:} {Baseline Task Accuracy} and \textbf{Right:} {Baseline Task Accuracy controlling for model size}. 
    Markers and error bars show means and standard deviations on both axes for LCEs.
    Both settings show the non-significant correlations.  Improvements in baseline task accuracy brought by model scales are insufficient to contribute to better LCE performance gains.}
    % \jr{need to add Appworld after finishing training}}
    \label{fig:baseacc}
    \vspace{-0.2in}
\end{figure}

\rev{In addition to failure-mode diversity, baseline task accuracy represents another potential contributor to the different performance gains. 
Here, baseline task accuracy is defined as the initial performance achieved using each LCE's baseline prompt on a specific task.
Specifically, we use the results from our scaling experiments, including 6 models from Qwen3-0.6B to 235B, 2 LCEs, and 4 datasets. For a more comprehensive understanding, we analyze the association of LCE performance gains with baseline task accuracy (\textbf{Left} in~\cref{fig:baseacc}) and with baseline task accuracy controlling for model size (\textbf{Right} in~\cref{fig:baseacc}).
We observe that neither analysis reveals a significant correlation.
These observations cannot provide evidence that higher baseline task accuracy reliably yields more LCE performance gains.}

\section{Evaluation with enhanced answer extraction}
\label{app:reeval}

\rev{
In~\cref{fig:lce_math500}, we observe that optimized contexts improve instruction adherence on Gemma4. Here we further investigate whether these improvements remain after normalizing answer formats. Specifically, we instruct Qwen3.8-27B to extract the final answer from the failures of Gemma4 to ensure better answer extraction (see the prompt in~\cref{fig:llmans}). As shown in~\cref{tab:reeval}, applying the LLM-based answer extractor substantially reduces the performance gains achieved by LCEs, suggesting that the original LCE performance gains mainly come from improved format adherence.
}

\begingroup
\captionsetup[figure]{position=bottom, skip=4pt}
\begin{lstlisting}[style=promptstyle, belowskip=0pt]
Extract the final answer from the response.
If no final answer can be identified, return an empty string.

Return exactly one JSON object:
{"answer": "extracted final answer"}
\end{lstlisting}
\captionof{figure}{Prompt Template of LLM-based Answer Extraction.}
\label{fig:llmans}
\endgroup

\begin{table}[h]
    \centering
    \small
    \caption{Performance of LCEs with and without the LLM-based answer extractor. The performance gains substantially decrease and nearly vanish after applying the extractor, indicating that the observed improvements achieved by LCEs stem primarily from better adherence to format instructions.}
    \label{tab:reeval}
    \begin{tabular}{l|ccc|ccc}
        \toprule
        \multirow{2}{*}{\textbf{Method}}
        & \multicolumn{3}{c|}{\textbf{w/ LLM-Extractor}}
        & \multicolumn{3}{c}{\textbf{w/o LLM-Extractor}} \\
        \cmidrule(lr){2-4} \cmidrule(lr){5-7}
        & Initial & Final & Gain 
        & Initial & Final & Gain \\
        \midrule
        GEPA    & 91.70 & 91.20 & -0.50 & 67.70 & 71.20 & +3.50 \\
        TF-GRPO & 91.10 & 92.60 & +1.50 & 67.70 & 73.80 & +6.10 \\
        ACE     & 88.80 & 88.00 & -0.80 & 67.80 & 69.40 & +1.60 \\
        \bottomrule
    \end{tabular}
\end{table}

\end{document}